\documentclass[preprint,12pt]{elsarticle}

\usepackage{amssymb}
\usepackage{amsmath}
\usepackage{graphicx}%
\usepackage{multirow}%
\usepackage{amsmath,amssymb,amsfonts}%
\usepackage{amsthm}%
\usepackage{mathrsfs}%
\usepackage[title]{appendix}%
\usepackage{xcolor}%
\usepackage{textcomp}%
\usepackage{manyfoot}%
\usepackage{booktabs}%
\usepackage{algorithm}%
\usepackage{algorithmicx}%
\usepackage{algpseudocode}%
\usepackage{listings}%
\usepackage{multirow}
\usepackage{comment}
\usepackage{enumitem}
\usepackage{tabularx}

\newcommand{\nr}{n^{\text{rob}}}
\newcommand{\Mprob}[1]{\mathcal{M}^{\text{p}}_{#1}}
\newcommand{\Mc}{\mathcal{M}^{\text{c}}_{k}}
\newcommand\card[1]{\left\lvert#1\right\rvert}
\newcommand{\st}[1]{s^{\text{r}}_{#1}}
\newcommand{\ste}[1]{s_{#1}}
\newcommand{\belief}[2]{b_{c,#1,#2}}
\newcommand{\obs}[2]{o_{#1,#2}}
\newcommand{\bzero}[1]{\belief{#1}{0}}
\newcommand{\nx}{n^{\text{h}}}
\newcommand{\ny}{n^{\text{v}}}
\newcommand{\xih}{\xi^{\text{h}}}
\newcommand{\xiv}{\xi^{\text{v}}}
\newcommand{\ph}{\lambda^{\text{h}}}
\newcommand{\pv}{\lambda^{\text{v}}}
\newcommand{\nF}{n^{\text{f}}}
\newcommand{\tilM}{\Tilde{\mathcal{M}}^{\text{env}}_{k}}
\newcommand{\tilvec}[2]{\Tilde{\boldsymbol{s}}_{#1,#2}}
\newcommand{\np}{n^{\text{p}}}
\newcommand{\nG}{n^{\text{goal}}}
\newcommand{\nC}{n^{\text{con}}}
\newcommand{\npath}{n^{\text{path}}}
\newcommand{\ntr}{n^{\text{travel}}}
\newcommand{\wG}{w^{\text{goal}}}

\newcommand{\wGP}{w^{\text{goal,P}}}
\newcommand{\wC}{w^{\text{con}}}

\newcommand{\whlC}{w^{\text{HL,C}}}
\newcommand{\wO}{w^{\text{agg}}}

\newcommand{\muG}{\mu^{\text{goal}}}
\newcommand{\muC}{\mu^{\text{con}}}
\newcommand{\muO}{\mu^{\text{agg}}}
\newcommand{\fv}[1]{\Tilde{s}_{#1}}
\newcommand{\wcom}[1]{w^{\text{tune}}_{#1}}
\newcommand{\wcomhl}[1]{w^{\text{tune,P}}_{#1}}
\newcommand{\wcomll}[1]{w^{\text{tune,C}}_{#1}}
\newcommand{\wcomup}[1]{\Bar{w}^{\text{tune}}_{#1}}
\newcommand{\dmax}{\delta^{\max}}

\newcommand{\Nxhat}[1]{\mathcal{X}_{#1}(\boldsymbol{x}_{#1})}
\newcommand{\elementX}[1]{\hat{\mathcal{X}}_{#1}(\hat{\boldsymbol{x}}_{#1})}

\newcommand{\kc}[1]{\kappa_{c,#1}}
\newcommand{\kj}[1]{\kappa_{j,#1}}
\newcommand{\mucluster}[2]{\mu^{\text{cluster}}_{#1}(#2)}
\newcommand{\ncluster}[1]{n^{\text{cluster}}_{#1}}
\newcommand{\score}[1]{s^{\text{cluster}}_{#1}}
\newcommand\norm[1]{\left\lVert#1\right\rVert}
\newcommand{\sexp}{s^{\text{exp}}}
\newcommand{\sunexp}{s^{\text{unexp}}}
\newcommand{\envir}[1]{\boldsymbol{e}_{#1}}

\theoremstyle{thmstyleone}%
\theoremstyle{thmstyletwo}%
\newtheorem{remark}{Remark}%

\theoremstyle{thmstylethree}%

\journal{Applied Soft Computing}

\begin{document}

\begin{frontmatter}

\title{Fuzzy-Logic-based model predictive control: A paradigm integrating optimal and common-sense  decision making}
 \author{Filip Surma\corref{cor1}\fnref{delftlabel}}
 \author{Anahita Jamshidnejad\fnref{delftlabel}}
\cortext[cor1]{Corresponding author.\\ E-mail addresses: f.surma@tudelft.nl (F. Surma), a.jamshidnejad@tudelft.nl (A. Jamshidnejad).}


\affiliation[delftlabel]{organization={TU Delft},
            addressline={Kluyverweg 1}, 
            city={Delft},
            postcode={2629 HS}, 
            state={South Holland},
            country={The Netherlands}}

\begin{abstract}
This paper introduces a novel concept, fuzzy-logic-based model predictive control (FLMPC), 
along with a multi-robot control approach for exploring unknown environments and locating targets. 
Traditional model predictive control (MPC) methods rely on Bayesian theory to represent environmental knowledge and optimize a stochastic cost function, often leading to high computational costs and lack of effectiveness in locating all the targets. 
Our approach instead leverages FLMPC and extends it to a bi-level parent-child architecture 
for enhanced coordination and extended decision making horizon.  
Extracting high-level information from probability distributions and local observations, 
FLMPC simplifies the optimization problem and significantly extends its operational horizon 
compared to other MPC methods.  
We conducted extensive simulations in unknown 2-dimensional environments 
with randomly placed obstacles and humans. 
We compared the performance and computation time of 
FLMPC against MPC with a stochastic cost function, then evaluated the impact of integrating 
the high-level parent FLMPC layer. 
The results indicate that our approaches significantly improve 
both performance and computation time, enhancing coordination of robots and 
reducing the impact of uncertainty in large-scale search and rescue environments.
\end{abstract}



\end{frontmatter}




\section{Introduction}

Since 2021, there have been over 200,000 deaths due to various natural and technological disasters \cite{disaster_website}. 
This number would even be larger without proper intervention of the rescue teams, showing the importance of systematic and rapid search and rescue in the aftermath of disasters. 
In urban search and rescue (USaR), a team of trained rescuers is dispatched to investigate unfamiliar environments and to locate and assist the trapped victims.  
To achieve an optimal and expedient completion of the USaR mission, it is imperative that the mission 
is operated time-efficiently, as prolonged search times are associated with a diminished probability of survival for the trapped victims  \cite{LocationNavigationSurvey,longMissionTimes}. 

In the modern era, USaR missions that used to be conducted solely by human rescuers, 
are increasingly delegated to teams composed of autonomous robots and humans, 
where the robots should take over tasks that are life-threatening for human rescuers, including the exploration of unknown environments \cite{humanRobotTeam}. 
Robots have been deployed in large-scale disasters, e.g., in the search and rescue operations for the 
tsunami that struck the Fukushima nuclear power plant in 2011 \cite{Fukushima}. 
Various types of robots  are currently used in search and rescue operations, 
including aerial robots \cite{UAV_USARsurvey} and (under-)ground robots \cite{icarus}, e.g., snake robots \cite{snakeRobotsMexico} and legged robots \cite{darpa}.%

The long-term objective of USaR is to use fully autonomous teams of robots in high-risk missions, 
without any need for human intervention. 
This is motivated by the inherent dangers associated with navigating damaged buildings. 
In such cases, firefighters may have a limited window of opportunity to evacuate.  
They should locate the exit expeditiously, while retreating through the same route they entered the building 
may not be possible due to ongoing structural damages \cite{LocationNavigationSurvey}. 
To make multi-robot teams effective in USaR missions, their decisions/actions should be coordinated \cite{fuzzyMPCsearch}. 
Crucially, these robots should be capable of making both short-term and long-term plans that ensure 
fast and safe exploration and evacuation of the building, despite uncertainties \cite{LocationNavigationSurvey}.%

\subsection{Solution proposed in this paper}

Model predictive control (MPC) is a widely known control method that has proven 
effective in various applications \cite{MPCbook}. 
MPC employs a model of the controlled system in order to predict the future states of the system 
under the impact of candidate MPC inputs. 
An optimization solver is used to identify a state trajectory that is optimal within a given prediction window, with regards to a given objective function and the state and control constraints. 
Despite its effectiveness, MPC struggles to solve large-scale control problems with multiple uncertainty levels.%

In this paper, we introduce a novel version of MPC, called 
fuzzy-logic-based model predictive control (FLMPC), which is then extended into a bi-level architecture 
that particularly suits spatially and/or temporally large-scale 
control problems that involve multiple levels of uncertainties, 
e.g., exploration of unknown environments through search and rescue robotics.  
We showcase the effectiveness of FLMPC, both in its original formulation and when re-formulated within the bi-level architecture, for multi-robot exploration mission planning via computer-based simulations.%

Our main motivation for introducing FLMPC is the following: 
As is later on in Section~\ref{sec:related} presented, a variety of approaches 
that base their decision making in uncertain situations on the use of probability values, 
face a common issue, i.e., they require to frequently update the probability distributions inside the 
control loop. This not only diminishes the efficiency of the solver, but also demands assumptions regarding all future states and measurements of the system \cite{motionEncodedPS,priorityEncoding}. 
In complex decision making tasks, e.g., in searching for targets in unknown areas, human experts prove very adept,  
while they do not focus on calculating any precise changes in the probabilities over time. 
Instead, humans rely on their “common sense” and adhere to fundamental principles such as “avoid danger” 
or “explore less known places first”. 
To emulate such an approach via a mathematical control system, we integrate fuzzy logic, 
a mathematical approach that handles imprecision by mimicking human logic, 
into the MPC formulation. 
FLMPC allows to make decisions based on multiple objectives and constraints, such that these 
decisions are effective in the long term and in challenging, uncertain environments.


In actual USaR operations, individuals are coordinated by an incident commander, who analyzes 
the available information and directs the efforts of numerous responders simultaneously \cite{humanRobotTeam}. 
This allows multiple rescuers to be coordinated, and enables them to locally make reactive and rapid decisions based on newly acquired information during the search process. 
Inspired by this, we propose a bi-level architecture, where in the higher level an FLMPC system  
tunes the parameters of one or multiple low-level FLMPC systems to steer them in relevant sub-sets of the admissible solutions, while these low-level FLMPC systems react properly to unexpected/unpredicted measurements, akin to human rescuers.%

\subsection{Main contribution and structure of the paper}

The main contributions of this paper are:
\begin{itemize}
\item 
\textbf{Developing FLMPC for effective handling of uncertainties:}
A novel fuzzy-logic-based model predictive control (FLMPC) framework is 
introduced for exploring unknown environments and for locating targets using dynamic fuzzy maps. 
FLMPC addresses the computational and performance limitations of traditional MPC approaches 
that rely on stochastic cost functions and Bayesian theory for interpreting unknown environments.
\item 
\textbf{Introducing bi-level parent-child architecture for multi-robot control:} 
A bi-level parent-child architecture is proposed to enhance coordination and to extend the 
decision making horizon of FLMPC. This improves the ability of robots to handle 
large-scale environments and multiple levels of uncertainty. 
\item 
\textbf{Evaluation and comparison:}
Extensive simulations are performed, demonstrating the superiority of FLMPC 
over conventional MPC with a stochastic cost function. 
They reveal significant improvements in both performance and computation time, 
as well as enhanced coordination and robustness in large-scale search and rescue missions.
\end{itemize}

The paper is structured as follows. Section~\ref{sec:related} reviews  
alternative approaches from related literature,  
Section~\ref{sec:proposed_methods} gives the problem statement, the  theory of FLMPC, and 
its adoption and leverage via a bi-level parent-child architecture for search-and-rescue robotics. 
Section~\ref{sec:implmentation} presents the results of $2$ case studies, comparing FLMPC with MPC 
using a stochastic cost function, and evaluating single-level versus bi-level parent-child FLMPC. 
Finally, Section~\ref{sec:conclusion}  concludes the paper and outlines future research directions.%

\section{Related work}
\label{sec:related}

The concept of FLMPC introduced in this paper and its implementation for USaR are based on fuzzy optimization, 
MPC, and grid-based exploration of unknown environments. 
Thus, we provide below a background discussion on these topics.%

\subsection{Fuzzy decision making}

Fuzzy logic was first introduced in \cite{Fuzzy_introduction} for describing deterministic uncertainties 
using fuzzy sets. 
The fuzziness arises from imprecise boundaries in quantifying a concept, 
rather than from inherent randomness or unpredictability. 
Fuzzy sets may in general be seen as mathematical representations of vague linguistic terms 
used by humans that are not (precisely) quantified. 
For example, terms, such as {“high”, “hot”, and “tall” possess no clear delineation 
with, respectively, “low”, “cold”, and “short”}. 
Consequently, in fuzzy set theory, a membership function is associated to each fuzzy set 
that provides the extent that an element or a value (e.g., temperature $26^{\circ}$C) 
belongs to that fuzzy set (e.g., the set {“hot”}).%

In \cite{bel}, fuzzy logic has been leveraged for decision making, particularly 
to control a given system or agent within a fuzzy environment with possibly multiple fuzzy goals 
(i.e., goals that, instead of a clear or binary distinction between 
being achieved and not being achieved, allow for a range of satisfactory outcomes associated with 
different degrees of fulfillment) and 
fuzzy (also called soft) constraints (i.e., constraints that allow flexibility or degrees of satisfaction, 
rather than being strictly binary). 
Each goal and constraint is represented by a fuzzy set, which receives a state or control input 
of the controlled system/agent as input. 
The objective of the fuzzy decision making system is to provide a multi-stage process 
in the fuzzy environment that results in a sequence of control inputs, given the current state 
of the controlled system/agent. This sequence should result in an optimal outcome for the controlled system/agent 
with regards to the (fuzzy) goals. 
In order to estimate the final outcome, an aggregation operator may be used 
that exhibits both monotonicity and commutativity properties. 
Monotonicity implies that when one input of the aggregation operator is increased, then the output 
either increases or remains unchanged. 
Commutativity implies that the order of the inputs to the aggregation operator does not affect the outcome \cite{fuzzy_optimization_aggregation_property}.


Later on, the concept of decision making in fuzzy environments served as the foundation for fuzzy optimization 
\cite{zimmermanFuzzyProgramming}, which emulates the decision making processes of 
humans when confronted with situations involving competing objectives. 
In \cite{fuzzyMPC}, fuzzy MPC (FMPC) was introduced, where conventional model predictive control (MPC) is solved using fuzzy optimization. FMPC was shown to result in a steady state for the controlled system more rapidly than regular 
MPC, although the primary disadvantage of FMPC was its high computation time. 
The most notable distinction of FMPC in comparison with previous methods based on fuzzy optimization 
is that FMPC performs aggregation operations separately on the fuzzy goals and on the fuzzy constraints, 
whereas pervious fuzzy optimization methods treated the fuzzy goals and constraints similarly 
and aggregated them altogether. 
Thus, FMPC properly addresses multi-stage control problems that involve delayed goals (particularly rewards). 
In such cases, adopting state realizations that do not result in immediate rewards may still be beneficial
when it allows to later on reach states that will result in very high cumulative rewards. 
Since fuzzy constraints should prohibit the realization of the  
higher-risk states, 
such an impact on the overall reward will be obstructed when the goals and constraints are merged and treated the same.
The stability of FMPC was proved  
by imposing hard terminal constraints and by assuming the existence of a locally stabilizing terminal control law 
that replaces the FMPC strategy as soon as the system reaches the hard terminal set \cite{fuzzyMPCstability}. 
It has later on been shown that replacing the FMPC strategy with a terminal control law deems unnecessary 
by including additional terminal fuzzy goals for the states \cite{fuzzyMPCstabilityExtended}.%

Recently, fuzzy systems and MPC have been integrated, with the following motivations: 
to incorporate the dynamics 
of the external disturbances into the decision making of MPC through fuzzy models \cite{Surma2024};   
to represent the nonlinear dynamics of the controlled system via fuzzy models embedded in MPC \cite{fuzzyModels};  
to include fuzzy decision making and MPC in modular or hierarchical control architectures that collaboratively 
steer the controlled systems with complex dynamics towards desired objectives \cite{fuzzyMPCsearch,FuzzyCooperativeSecondaryControl}.%

\subsection{MPC in exploration of unknown environments via robots}
\label{sec:MPC_exploration}

Path planning for multiple robots that should explore unknown environments 
is a highly relevant problem for various applications, including USaR. 
Recently, methods based on MPC \cite{Surma2024,Baglioni2024}, approximate versions of 
MPC \cite{Surma2024_IFAC}, or MPC combined with fuzzy decision making \cite{fuzzyMPCsearch} 
have been proposed for path planning of robots that should explore USaR environments. 
MPC is gaining more attention in the field due to its unique capabilities in 
addressing multi-objective USaR missions with various (safety-related) constraints.%

It is common to formulate the exploration problem of the robots in the discrete space domain, 
by representing the environment via a grid, where each cell in the grid may be empty or 
occupied by various elements. 
After initializing their positions, at every time step the robots make either 
an omnidirectional (in all possible directions, e.g., forward, backward, and to the sides) or 
a restricted (e.g., only in specific directions, e.g., forward, forward-right, forward-left) 
movement that takes into account their limited maneuverability  \cite{motionEncodedPS,priorityEncoding,motionEncodedGA,distrbuitedCooperativeSearch,searchUAVlearning}. 
After each movement, the robots receive new measurements from the environment and update their 
map of the environment, commonly using the Bayesian rule \cite{motionEncodedGA,ProbabiliticRobotics}.

The mission planning for the robots is often represented as a multi-stage nonlinear non-convex optimization problem 
that generally includes both the constraints and the objective function of USaR in the decision making  \cite{motionEncodedPS,priorityEncoding,motionEncodedGA,distrbuitedCooperativeSearch,motionEncodedSCS,DMPCsearch}. 
However, solving such optimization problems faces two main challenges: 
First, since the robots continuously receive new information about their environment 
during the mission, the optimization problem should be re-solved very frequently \cite{DMPCsearch}. 
Second, due to the non-convex nature of the problem and the limited time to solve it in real-world USaR missions, 
the returned solutions, if received in time, are likely to be sub-optimal, thus degrading the overall performance. 
Therefore, the formulation of the optimization problem is of utmost importance and 
the decision variables should efficiently encode the most relevant information.%

Each MPC input, in general, is a single decision variable that is determined for all time steps 
along the prediction horizon. In exploring unknown environments that are represented by a grid through robots, 
the decision variable is commonly the movement of the robots per discrete time step, 
i.e., the direction or position of the next cell the robot should move to. 
This approach in literature is often referred to as motion encoding \cite{motionEncodedPS,priorityEncoding,motionEncodedGA,motionEncodedSCS}, 
which faces multiple challenges: 
Determining long paths for large-scale environments via motion encoding results in large 
numbers of decision variables, thus in excessive computational efforts. 
Moreover, since the selection of each candidate path depends on the position of all its cells, 
changing the value of one decision variable in motion encoding requires changing the entire path 
\cite{priorityEncoding}. This implies that the {value of the objective function} is very sensitive to single changes in the decision variables.  
Finally, in various cases the optimal paths are simply the shortest paths that connect the current position of the robot to its target via straight lines. 
In such cases, motion encoding results in unnecessarily large computational costs.
In \cite{priorityEncoding}, priority encoding has been proposed, where a number 
is initially decided, as a hyper-parameter, for the goals, which are the decision variables. 
The path is determined by connecting these goals through straight lines, leading 
to fewer decision variables and, thus, to reduce sensitivity of the {value of the objective function} to any changes in the decision variables.

In selecting optimal paths for robots, multiple candidate paths are usually graded 
and compared according to an objective function. Maximizing the probability of detecting the targets 
across the prediction horizon is 
a common objective in literature \cite{motionEncodedPS,motionEncodedGA}, but it does not guarantee the 
minimization of the detection time, which is also crucial in USaR \cite{antColony}. 
Alternatively, a discounted detection utility function has been proposed to formulate the objective 
function in \cite{BayesianOptimizationAglortihm}. Next to maximizing 
the cumulative probability of target detection, a discount factor encourages the 
robots to prioritize visiting locations that have a higher probability of detecting the targets, 
aiming at speeding up the target detection procedure.%

In order to incorporate the USaR uncertainties into the decision making, 
the path planning problem is usually formulated within a stochastic framework. 
However, solving the optimization problem within a stochastic framework using MPC results 
in complex representations for the objective function, for instance, the cumulative expected value 
of an objective within the prediction horizon. 
Predicting the probability distribution for all the cells 
in real life, however, result{s} in expensive computations. 
Alternatively, simplifying assumptions may be made about the future measurements in the cells: 
For instance, in \cite{priorityEncoding} the cells that {are} detected within the prediction window 
are assumed to be vacant.%

In general, probability-based objective functions are suited mainly when knowledge about the 
initial probability distribution of the target locations in the environment exists. 
Otherwise, additional non-probabilistic terms should be included in the objective function, 
e.g., the (un)certainty level about the environmental knowledge, 
to encourage the robots to visit unexplored parts of the environment \cite{fuzzyMPCsearch,priorityEncoding}. 
A detection-based reward may also be included in the objective function 
to steer the robots to visit cells that include the targets \cite{priorityEncoding}. 
A pheromone map has been used in \cite{distrbuitedCooperativeSearch}, in order 
to enhance the coordination of the robots, by discouraging them from moving (close) to cells that are already visited. This prevents the robots from limiting their exploration to the same neighborhood.  
A motion cost may also be incorporated within the objective function, in order to discourage the robots from unnecessary movements \cite{priorityEncoding}.%

A crucial step in solving the path planning optimization problem is to choose a proper optimization solver (i.e., an algorithm that finds an optimal solution). 
The literature suggests, among others, the genetic algorithm \cite{priorityEncoding,motionEncodedGA}, 
particle swarm algorithm \cite{motionEncodedPS}, sand cat swarm \cite{motionEncodedSCS}, Bayesian optimization algorithm \cite{BayesianOptimizationAglortihm}, and ant colony optimization \cite{antColony}.
Additionally, it should be decided whether or not the optimization problem is 
formulated in a decentralized/distributed \cite{distrbuitedCooperativeSearch,DMPCsearch}, a centralized \cite{priorityEncoding}, 
or a combined framework \cite{motionEncodedGA}. 
For example, in \cite{motionEncodedGA} the targets (whose positions are modeled according to a given distribution) are divided into groups 
and each robot is assigned to one group of these targets and performs independently from other robots 
(since robots do not contribute to each other's decision making, this is {a} decentralized architecture). 
In \cite{fuzzyMPCsearch}, the robots use both a decentralized framework, based on fuzzy inference systems, and a centralized controller, based on MPC.%

\section{Proposed methods}
\label{sec:proposed_methods}

In this section, we explain the details of our proposed methods for 
autonomous multi-robot path planning in unknown and uncertain environments. 
This includes the problem statement, the new concept of fuzzy-logic-based 
model predictive control (FLMPC), leveraging FLMPC for exploration of unknown environments, 
and a novel bi-level architecture, called the parent-child framework, in which we formulate 
and embed FLMPC for efficient solution of complex large-scale path planning problems.%


\subsection{Problem statement}
\label{sec:problem_statement}

We consider $\nr$ number of autonomous robots (e.g., flying/ground search robots) that should map 
an initially unknown environment represented by a two-dimensional grid of $\nx$ horizontal and $\ny$ vertical cells, 
where all cells in the grid have the same size and shape (namely that of a square). 
The dynamics of the control system that is used to steer the robots  
in {the environment is} formulated in the discrete time. 
Assuming a static environment, each cell $c$ in the environment is represented via its 
horizontal and vertical coordinates $[\xih_c,\xiv_c]^\top$, where 
$\xih_c \in \{1, \ldots \nx\}$ and $\xiv_c \in \{ 1, \ldots, \ny \}$, and is associated with a state  $\st{c}$ 
that contains all the relevant information that describes the status of the cell.  
In particular, the state of a cell in a USaR mission may include information about its occupancy status  
(e.g., empty, embedding humans, occupied by obstacles, etc.).  
The admissible set of the different realizations for  $\st{c}$ is shown by $\mathcal{S}$. 
In general, this set is given as an ordered set $\mathcal{S} = \{\rho_1, \ldots, \rho_{\card{S}}\}$, where $\card{\cdot}$ is used for the cardinality of the set
and $\rho_i$'s for $i = 1, \ldots, \card{\mathcal{S}}$ are all possible realizations of $\st{c}$ for all cells $c$ in the environment. At every time step, robots can move to an adjacent cell (including diagonal movements). {Note that the states associated to the cells may correspond to different real-life concepts. For the robots to distinguish these concepts, the set $\mathcal{S}$ is defined to be ordered.}

While the environment is assumed to remain unchanged in time, the knowledge about the environment 
evolves in time as the robots explore the environment. 
Thus, per time step $k$ the path planning system has access to an estimate $\ste{c,k}$ of the real state $\st{c}$ of each cell in the environment, where this estimation is updated in time. 
No data or information is initially available regarding the content of the cells or 
the number of the humans in the environment. 
As the robots traverse the environment, they scan it via their sensors, which in general 
may be prone to imperfections. The objective of the multi-robot system is to detect/rescue as  
many victims as possible in the shortest possible time, while reducing the number of the robots 
that become out of function (e.g., that are crashed into or that are caught in fire) during the search and rescue.

Table~\ref{tab:notations} shows the mathematical notations that are frequently used in the paper, together with their definitions. We use a regular font for scalars and a bold font for vectors.%

\begin{table}[]\tiny\caption{Frequently used mathematical notations}\label{tab:notations}
\begin{tabular}{ll}
$b_{c,i,k}$     & Probability of a cell $c$ having $\rho_i$ as a realization of a state  at time-step $k$  \\
$c$                               & 2-element vector containing position of a cell                                                                                    \\
$\envir{k}$                          & Current knowledge about the environment                                                 \\
$f(\cdot)$                        & Model of system's dynamics                                                              \\
$\Tilde{f}_l(\cdot)$                      & Model describing how to update fuzzy maps                                               \\
$\mathcal{M}^{\text{c}}_k$           & Certainty map   at time-step $k$                                                     \\
$\Tilde{\mathcal{M}^{\text{env}}}_k$  & Set of all fuzzy maps     at time-step $k$                                                                 \\
$\mathcal{M}^{\text{p}}_k$           & Probability map    at time-step $k$                                                \\
$\ncluster{k}$                   & Number of clusters                                                                      \\
$n^{\text{con}}$                             & Number of constraints                                                            \\
$n^{\text{f}}$                  & Number of fuzzy states required to describe a cell                                     \\
$n^{\text{goal}}$   
& Number of goals        \\
$n^{\text{h}}$                    & Horizontal size of a grid                                                                           \\
$n^{\text{p}}$                              & Prediction horizon (for MPC in general)                                                 \\
$n^{\text{path}}$                    & The longest possible path computed during optimization                                  \\
$n^{\text{rob}}$                  & Number of robots                                                                        \\
$n^{\text{travel}}$             & The longest possible part of a path defined by a single decision variable                                  \\
$n^{\text{v}}$                    & Vertical size of a grid                                                                 \\
$o_c(k)$                          & Observation of cell $c$ at time-step $k$                                                \\
$\mathcal{O}(k)$                  & Set of all measurements at time-step $k$                                                \\
$p(\cdot|\cdot)$                 & Conditional probability                                                                 \\
$\mathbb{P}$                     & An ordered set of cells to be visited by a robot computed by parent FLMPC                             \\
$r$                               & Reference                                                                               \\
$\mathcal{R}$                    & Set of all robots                                                                      \\
$\boldsymbol{\Tilde{s}}_{c,k}$              & A vector of fuzzy states describing a cell $c$ at time-step $k$                           \\
$\Tilde{s}_{c,l,k}$              & A $l^{\text{th}}$ fuzzy state of a cell $c$ at time-step $k$                           \\
$s_{c,k}$                             & An estimate of a state of a cell $c$                                                  \\
$\score{j,k}$                    & A score given to the $j^{\text{th}}$ cluster at the time-step $k$                      \\
$\sexp$                          & If score of a cluster is below this value, it is considered explored.                  \\
$s^\text{r}_c$                    & A real state of a cell $c$                                                             \\
$\sunexp$                          & If score of a cluster is above this value, it is considered unexplored.                  \\
$\mathcal{S}$                    & The admissible set of all possible states of a cell in the environment.                                              \\
$\boldsymbol{u}_k$                & Input at time step $k$                                                                       \\
$w^{\text{agg}}$                             & Parameter used in the final aggregation                                                 \\
$w^{\text{con}}$                             & Parameter used to aggregate constraints                                                 \\
$w^{\text{goal}}$                             & Parameter used to aggregate goals                                                       \\
$\wGP$                           & Parameter used to aggregate goals but used by Parent FLMPC                               \\
$\whlC(c)$                   & A weight given to a cell by long-term FLMPC                                             \\
$w^{\text{cluster}}$           &   A weight used in computing scores for clusters                                                                \\
$\wcom{c,\kc{k}}$                  & A parameter that shows the weight given to a cell $c$ at time-step $k$ used by single level FLMPC                \\
$\wcomll{j,\kj{k}}$               & A parameter that shows the weight given to a cell $c$ at time-step $k$ used by Child FLMPC                \\
$\wcomhl{j,\kj{k}}$               & A parameter that shows the weight given to a cell $c$ at time-step $k$ used by Parent FLMPC                \\
$\wcom{c,\kc{k},r}$        & A parameter that shows the weight given to a cell $c$ at time-step $k$  by the $r^{\text{th}}$ robot       \\
$\wcomup{c,\kc{k},r}$     & A parameter that shows the weight given to a cell $c$ at time-step $k$  by the $r^{\text{th}}$ \\ & robot while taking into consideration other robots' weights \\
$\boldsymbol{x}_k$                & State of the system at time $k$                                                                       \\
$\boldsymbol{\hat{x}}_k$          & Trajectory of states starting at time $k$                                               \\
$\boldsymbol{\hat{x}}_k^{\text{neigh}}$       & All states in the neighborhood of all states in trajectory $\boldsymbol{\hat{x}}_k$    \\
$\mathcal{X}_k$           & Set of all observable states along a trajectory $\boldsymbol{\hat{x}}_k$  and weights allocated to them    \\
$\mathcal{X}_k^r$         & Set of all robots' $\mathcal{X}_k$                                                             \\
$z_c(k)$                          & Certainty of cell $c$ at time-step $k$                                                  \\
$\alpha(\cdot)$                   & A parameter that shows how distance affects goals                                       \\
$\beta(\cdot)$                    & A parameter that shows how time affects goals                                           \\
$\gamma$                          & A constant disquant factor between 0 and 1 used to show how time affects cost                      \\
$\delta_{c,\kc{k}}$              &  Predicted distance between a robot and a cell $c$ at time step $\kappa$               \\
$\delta^{\max}$                  & Maximum distance sensed around a robot                                          \\
$\kappa$                         & This parameter represents the predicted time-step  \\
$\lambda_{i,k}^{\text{h}}$        & Horizontal position of a center of the $i^{\text{th}}$ cluster at time step $k$        \\
$\lambda_{i,k}^{\text{v}}$        & Vertical position of a center of the $i^{\text{th}}$ cluster at time step $k$        \\
$\mu^{\text{agg}}(\cdot)$                    & Function that returns how a trajectory of states satisfies all goals and constraints    \\
$\mucluster{i,k}{c}$             & Membership of a cell belonging to the $i^{\text{th}}$ cluster at time step $k$                 \\ 
$\mu^{\text{con}}_i(\cdot)$                  & Function that returns how a state satisfy $i^{\text{th}}$ constraint                    \\
$\mu^{\text{con}}(\cdot)$                    & Function that returns how a trajectory of states satisfies all constraints              \\
$\mu^{\text{goal}}_i(\cdot)$                  & Function that returns how a state satisfy $i^{\text{th}}$ goal                          \\
$\mu^{\text{goal}}(\cdot)$                    & Function that returns how a trajectory of states satisfies all goals                    \\
$\xi^{\text{h}}_c$                &  Horizontal coordinate of a cell                                                       \\
$\xi^{\text{v}}_c$                &  Vertical coordinate of a cell                                                         \\
$\rho_i$                          & $i^{\text{th}}$ possible state                                                          \\
$\sigma$                          & A constant parameter used to compute $\whlC(c)$                                          
\end{tabular}
\end{table}

\subsection{Fuzzy-logic-based model predictive control (FLMPC)}
\label{sec:FLMPC}

This section introduces fuzzy-logic-based model predictive control (FLMPC), where the discussions are based on the following assumptions: 
\smallskip
\begin{enumerate}[label=\textbf{A\arabic*}]
\item \label{ass:zero}
FLMPC determines optimal trajectories for a system, which then {follows} these trajectories to dynamically interact with an initially unknown environment. 
\item  \label{ass:one}
There is a finite number of admissible states for the environment and these are known to the system.
\item  \label{ass:two}
The states of the cells are generally graded according to multiple, potentially competing objectives. 
\item \label{ass:three}
The system is generally exposed to constraints, which depend on the environmental states, and that should be satisfied when interacting with the environment. 
\end{enumerate}
\bigskip

FLMPC is introduced for a system that should dynamically interact with an environment prone to Assumption~\ref{ass:one}.
Due to the environmental uncertainties, the objectives of the system (prone to Assumption~\ref{ass:two}) and its constraints (prone to Assumption~\ref{ass:three}) are described as fuzzy goals and fuzzy constraints, 
mathematically given by fuzzy sets and, thus, associated with membership functions. 
Each membership function, thus, receives as input the current state $\boldsymbol{x}_k$ of the system and the current knowledge  $\envir{k}$ 
about the states of the environment (a vector that includes $\ste{c,k}$ at time step $k$ for all cells of the environment), 
and returns a value in $[0 , 1]$ that reflects the importance of the state with regards to that {fuzzy} goal. 
This, for the fuzzy constraints, implies that the realization of those states whose membership degree to the 
corresponding fuzzy sets is less than 1 penalizes the system.%

FLMPC determines a candidate state trajectory $\hat{ \boldsymbol{x}}_k=\{\boldsymbol{x}_{k+1}, \boldsymbol{x}_{k+2}, \dots, \boldsymbol{x}_{k+ \np}\}$, 
based on the candidate input trajectory $\hat{\boldsymbol{u}}_k=\{\boldsymbol{u}_{k},\boldsymbol{u}_{k+1},\dots,\boldsymbol{u}_{k+ \np - 1}\}$, for the system 
 per time step $k$, with the hyper-parameter $\np$ representing the prediction horizon. 
 This state trajectory should maximize the consensus for all different objectives, 
 violating {or being penalized by} less constraints. 
Each candidate trajectory involves a total number of $\np (\nG + \nC)$ 
importance/satisfaction degrees, with $\nG$ and $\nC$ the total number of {fuzzy goals and fuzzy} constraints, respectively. 
Therefore, for a solver to identify an optimal solution, a single aggregated degree 
of importance should be obtained to be used to evaluate the entire trajectory.  
In fuzzy optimization, where the goals and the constraints are represented via fuzzy sets, aggregation operations are employed 
at the end, {after estimating individual degrees of importance,} to derive the overall degree of importance for a given solution. 
This allows to take into account the satisfaction levels associated with both the goals and the constraints. 
For FLMPC, we perform three primary operations: (1) aggregation of the importance levels of a candidate solution with regard to all the fuzzy goals; (2) aggregation of the satisfaction levels for all the fuzzy constraints;  
(3) aggregation of the results from the previous two stages to determine the best solution at the final stage.%

\subsubsection{Aggregating the importance degrees for all fuzzy goals}
\label{sec:agg_goals}

Suppose that the candidate solution of FLMPC at time step $k$ is trajectory $\hat{\boldsymbol{x}}_k$. 
Note that we assume there is no prediction model that provides information about the future 
perception of the environment, and thus, for the sake of simplicity we always rely on the 
knowledge $\envir{k}$ {available at time step $k$ within the entire} prediction horizon $\{k+1, \ldots, k+\np\}$. 
In order to aggregate the degrees of importance with regard to all the goals 
and to determine an overall degree of importance $\muG\left(\hat{\boldsymbol{x}}_k,\envir{k}\right)$ 
for this trajectory, we use the generalized mean, given below: 
\begin{equation}
\label{eq:mean}
\muG\left(\hat{\boldsymbol{x}}_k,\envir{k}\right) = \left(\frac{1}{\np}\sum_{j = 1} ^{\np}
\Big(\muG_1\left({\boldsymbol{x}}_{k+j},\envir{k}\right) \star \ldots \star \muG_{\nG}\left({\boldsymbol{x}}_{k+j},\envir{k}\right)\Big)^{\wG} \right)^\frac{1}{\wG}
\end{equation}
with $\muG_i\left({\boldsymbol{x}}_{k+j},\envir{k}\right)$ for $i \in \{1, \ldots,\nG\}$ the degree of importance associated to state {$\boldsymbol{x}_{k+j}$} with regards to the $i^{\text{th}}$ goal, based on the current knowledge $\envir{k}$ about the states of the environment, and $\star$ the symbol used for an S-norm \cite{Fuzzy_introduction}. 
Depending on the application, obtaining a larger number of high-valued rewards may excel 
achieving a constant stream of low-valued rewards, even though both cases may result in the same 
mean value. 
For instance, detecting a human (associated with a single high-valued reward) 
may generally be more important than exploring a larger part of the environment (associated with 
a stream of low-valued rewards). 
Tuning the weight $\wG$ in \eqref{eq:mean} allows to distinguish the relative importance 
of such candidate trajectories.  
A larger value for $\wG$ may be considered to give more power to importance degrees that are larger (i.e., are closer to $1$), 
resulting in a generalized mean that is closer to the maximum importance degrees.%


\subsubsection{Aggregating the satisfaction degrees for all fuzzy constraints}
\label{sec:agg_constraints}

In safety-critical applications, e.g., in USaR, those constraints whose 
violation is at higher risks, i.e., correspond to the least membership degrees for the given solution trajectory, 
play a crucial role in the decision making. 
Nevertheless, using the minimum function to aggregate the satisfaction levels of all the constraints poses two 
problems: {First, the minimum function treats two trajectories equally if their worst constraint violations have the same membership degree, even though the trajectory with more violations  should be less preferred.
Second, it is usually very} challenging for an optimization solver to identify an optimal state trajectory when a single state 
(the one associated to the highest risk) within the entire state trajectory 
evaluates the constraint satisfaction. %
Therefore, we propose to use the parameterized Yager T-norm function given below to estimate the 
overall degree $\muC\left(\hat{\boldsymbol{x}}_k,\envir{k}\right)$ of satisfaction {of the fuzzy constraints for the candidate 
state trajectory $\hat{\boldsymbol{x}}_k$. We have:}
\begin{equation}
\label{eq:yager}
    \muC\left(\hat{\boldsymbol{x}}_k,\envir{k}\right) = \max\left\{0,1-\left(\sum_{j = 1}^{\np}\sum_{i = 1}^{\nC} \left(1-\Big(\muC_i\left({\boldsymbol{x}}_{k+j},\envir{k}\right)\Big)^{\wC}\right)\right)^\frac{1}{\wC}\right\}
\end{equation}
with $\muC_i\left({\boldsymbol{x}}_{k+j},\envir{k}\right)$ the degree of satisfaction of the 
$i^{\text{th}}$ constraint by state {$\boldsymbol{x}_{k+j}$} within the candidate trajectory 
$\hat{\boldsymbol{x}}_k$, given the current knowledge $\envir{k}$ about the states of the environment.  
Unlike the minimum operator, this aggregation operator considers the satisfaction levels 
corresponding to states within a candidate trajectory.  
Moreover, for $\wC > 1$ it returns an aggregated value that is smaller than the 
aggregated value determined by the minimum operator for the same set of input values. 
This is because \eqref{eq:yager} introduces an additional penalty for all other constraints whose satisfaction degree 
is below 1. The weight $\wC$ is calibrated to tune this penalty properly.%

\subsubsection{Overall aggregation for determining the solution of FLMPC}
\label{sec:agg_overall}

Ultimately, in order to associate an overall membership degree $\muO\left(\hat{\boldsymbol{x}}_k,\envir{k}\right)$ 
to the candidate state trajectory $\hat{\boldsymbol{x}}_k$, the following function is proposed for FLMPC:  
\begin{equation}
\label{eq:multiplication}
     \muO\left(\hat{\boldsymbol{x}}_k,\envir{k}\right) = \muG\left(\hat{\boldsymbol{x}}_k,\envir{k}\right) * \left(\muC\left(\hat{\boldsymbol{x}}_k,\envir{k}\right)\right)^{\wO}
\end{equation}
where the symbol $*$ represents a standard T-norm in fuzzy logic \cite{Fuzzy_introduction}. 
The positive weight $\wO$ specifies the relative importance between the fuzzy goals and the fuzzy constraints. For $\wO\leq 1$, the goals are deemed to be of greater importance than the constraints, and vice versa.
\bigskip

\subsubsection{{Optimization problem of FLMPC}}

The following constrained optimization problem is solved by FLMPC, in order to determine an 
optimal state trajectory $\hat{\boldsymbol{x}}_k$ that corresponds to the maximum overall aggregated grade:  
\begin{subequations}
\label{eq:optimization}
\begin{align}
\begin{split}\label{eq:cost_func_general}
    \max_{\hat{\boldsymbol{x}}_k}\muO\left(\hat{\boldsymbol{x}}_k,\envir{k}\right)
\end{split}
\\
&\text{such that for }\kappa \in \{ k + 1, \ldots, k+\np \}: 
\nonumber\\
\begin{split}
\label{eq:dynamic_update_FLMPC}
    \boldsymbol{x}_{\kappa+1}=f(\boldsymbol{x}_{\kappa},\boldsymbol{u}_{\kappa})
\end{split}
\\
\begin{split}
    \hat{\boldsymbol{x}}_k=\{\boldsymbol{x}_{k+1},\dots,\boldsymbol{x}_{k+ \np}\}
\end{split}
\end{align}
\end{subequations}
where the constraint given by \eqref{eq:dynamic_update_FLMPC} describes the dynamics of the system, i.e., 
the evolution of its states based on the given inputs.%

In general, the constrained optimization problem \eqref{eq:optimization} is nonlinear and non-convex.


%
\smallskip


\begin{remark}
The aggregation operations given in \eqref{eq:mean} and \eqref{eq:yager} are examples that have 
been chosen for their {generality and} useful properties (e.g., the ability to be fine-tuned by adjusting their hyper-parameters 
for the purposes of USaR). 
It is of course possible to implement FLMPC with different aggregation operators, depending on the application and the problem. 
\end{remark}
    %

\subsection{FLMPC for exploring unknown environments}
\label{sec:FLMPC_USaR}

This section explains how FLMPC is adopted to solve the problem described in Section~\ref{sec:problem_statement}, where Assumptions~\ref{ass:one} and \ref{ass:two} hold, i.e., the environment has a limited number of cells 
in a grid map with a finite number of states per cell, and the possible control inputs are constrained what means that there is a finite number of possible actions and states. Moreover, multiple objectives are possible. 
Additionally, the following assumptions are considered:
\smallskip
\begin{enumerate}[label=\textbf{A\arabic*}]
\setcounter{enumi}{4}
\item \label{ass:five}
Each time a robot visits a cell, it collects observations from that cell and all cells in a given neighborhood 
of the cell, i.e., a circular area of radius $\dmax$.
\item \label{ass:six}
The certainty about the observations for the cells  that are in its  neighborhood 
declines with the distance of the cell from the robot.  
\item \label{ass:seven}
There is a finite number of candidate trajectories per time step per robot 
that are assessed by FLMPC to determine an optimal one. 
\end{enumerate}
%
Next, we discuss how to grade various candidate trajectories for robots, given 
maps that capture and are updated based on environmental information. 
This step is needed, in order to solve the FLMPC optimization problem \eqref{eq:optimization}
for mission planning of search and rescue robots.%

\subsubsection{Mapping unknown environments}
\label{sec:mapping_environment}

While the states of the cells in an unknown environment are not (fully) known, 
per time step the most recent data should be used to update the perception of the robots 
about these states for planning or updating the mission of the robots. 
One prevalent method for organizing the knowledge of the environment is through the creation of maps 
that {are} regularly updated when new data or information is captured from the environment. 

Probability and certainty maps, explained {next, are the most common in the literature.}
Research shows that integration of probabilistic and fuzzy knowledge about uncertain environments 
significantly enhances the environmental awareness of autonomous {robots} \cite{CEE:2024}. 
We accordingly add to the above maps, the concept of fuzzy maps that is particularly useful for autonomous USaR via robots in unknown and uncertain environments. Note that in the mathematical notations, fuzzy concepts are distinguished from 
non-fuzzy concepts via a tilde symbol on top of the fuzzy notations.%

\paragraph{Probability maps}
\label{sec:probability_map}

A probability map $\Mprob{k}$ at time step $k$ contains the probability distribution for all possible realizations of the states per cell for the given time step. 
The size of the probability map is, thus, $\nx \times \ny  \times \card{\mathcal{S}}$, i.e., 
it may be considered as a matrix of size $\nx \times \ny $ (i.e., the total number 
of cells in the environment), for which each element is a vector of 
size $\card{\mathcal{S}}$: The vector in position $\left(\xih_c, \xiv_c \right)$ 
of the probability map matrix includes the probability distribution for state {variable $\ste{c,k}$} 
of cell $c$ in the environment, with the $i^{\text{th}}$ element of this vector  showing the probability, estimated at time step $k$,  
that the realized value of state {$\ste{c,k}$} is $\rho_i$ (i.e., the $i^{\text{th}}$ element of set $\mathcal{S}$, as defined in Section~\ref{sec:problem_statement}). 
This element of the vector is called the {“belief”} that the realization of the state of cell $c$
at time step $k$ is $\rho_i$ and is denoted by $\belief{i}{k}$.%

The beliefs about the states of the cells are, in general, subject to changes over time. 
Prior to the start of the mission, an initial probability distribution, including initial values 
$\bzero{i}$ for $i \in \{1, \ldots, \card{\mathcal{S}}\}$, is assigned to all cells $c$ with $[\xih_c , \xiv_c]^\top \in \left\{ [1,1]^\top, \ldots, [\nx,\ny]^\top\right\}$. 
In the absence of initial environmental knowledge, the initial probability distributions may follow a uniform distribution for the cells. 
Alternatively, the values for $\bzero{i}$ may be initialized based on the knowledge about the environment before the disaster {occurred}.%

In the course of the exploration, the robots collect data via their sensors, 
e.g., in the form of measurements or images.  
Each measurement, which is also referred to as an observation, 
is specified by $\obs{c}{k}$ for cell $c$ at time step $k$. 
This data is then used to 
update the beliefs regarding the states of the cells in the environment. 
The following equation, inspired by the work in \cite{ProbabiliticRobotics}, 
is used to update the elements of the probability map for all cells $c$ with $[\xih_c , \xiv_c]^\top \in \left\{ [1,1]^\top, \ldots, [\nx,\ny]^\top\right\}$: 
\begin{equation}\label{eq:sensor_model}
    \belief{i}{k+1}=\frac{p\left(\obs{c}{k}|\rho_i\right)}
    {\sum^{\card{\mathcal{S}}}_{\ell=1}p\left(\obs{c}{k}|\rho_{\ell}\right)\belief{\ell}{k}} 
    \belief{i}{k}
    \qquad 
    \text{for $i = 1, \ldots, \card{S}$}
\end{equation}
where $p(\obs{c}{k}|\rho_i)$ is the likelihood that the observation in cell $c$ at time step $k$ 
is $\obs{c}{k}$, whereas the real state {$\st{c}$ of the cell is $\rho_i$. This likelihood may depend on the sensing capabilities or accuracy.}


\paragraph{Certainty maps}

In exploration via robotic systems probability maps are common, where current and/or future beliefs 
are incorporated into the objective function that should be optimized by the robots while exploring the environment. 
Nevertheless, alternative methods exist that employ other types of environmental maps. 
For instance, a prevalent map is the certainty map $\Mc$, a matrix of size $\nx \times \ny$ 
with time-varying elements in $[0,1]$, represented by $z_{c,k}$ for cell $c$ (i.e., for the element 
in position $\left(\xih_c, \xiv_c \right)$ of the certainty map) at time step $k$. 
Each value indicates the degree of certainty regarding the observation acquired for the corresponding cell.%

The certainty map is initialized with zeros, where these values increase compared to their values of the previous time step  with the acquisition of new observations. The certainty degrees remain unchanged or decrease (for dynamic environments) in time when no new observations are acquired.  
As is indicated in Assumptions~\ref{ass:five} and \ref{ass:six}, when visiting a cell the robot acquires observations from the neighboring cells too. 
The farther cell $c$ from the robot, the smaller the contribution of this observation to the enhancement of $z_{c,k}$.%




On the one hand, considering only one type of environmental maps may hinder the performance of the robotic system. 
On the other hand, considering multiple maps 
may be computationally unaffordable for real-time large-scale USaR robotic applications, due to the 
high costs of updating the maps inside the optimization loop of the robots. Thus, it is crucial to find a 
reliable, computationally efficient approach to model the environment, capturing as much relevant information as possible. This is the main motivation for proposing fuzzy environmental maps next.%

\paragraph{Fuzzy maps}

In this paper, we propose to model the perception of the robots about the environmental states via various low-level fuzzy maps, e.g., fuzzy certainty maps.  
These low-level fuzzy maps altogether build a matrix of size $\nx \times \ny$ that, instead of crisp variables, includes vectors of fuzzy variables as its elements. These fuzzy variables are each associated with one feature of the environment, e.g., the degree of certainty that a cell belongs to a specific category (represented as a label or fuzzy set) of certainty.%

The main advantage of fuzzy environmental maps, as {is} showcased in this paper, 
is that these maps are perfectly compatible with FLMPC and, when embedded in FLMPC, {
can be updated outside the optimization loop. This is because their fuzzy, high-level nature captures a broader range of information than probability maps, allowing for less frequent updates.} Thus, the resulting computational burden on online resources is significantly reduced. 
When an environment is modeled via $\nF$ low-level fuzzy maps, each cell 
of the environment is in fact described by a fuzzy vector of dimension $\nF$. 
Unlike probability maps, with low-level fuzzy maps there is more flexibility  in modeling 
various sources of uncertainty, in accordance with one's common sense \cite{fuzzyControl}.%

New observations are first used to update the probability map. 
Then, the elements of the low-level fuzzy maps are updated based on the new observations, the updated probability map, and the most recent fuzzy maps. 
Each element of fuzzy vector $\tilvec{c}{k} = [\fv{c,1,k}, \ldots, \fv{c,\nF,k}]^\top$, 
a vector that includes all the fuzzy characteristics/states associated to cell $c$ and that is placed at position $\left(\xih_c,\xiv_c\right)$ of the fuzzy map $\tilM$ of the environment at time step $k$, is updated by:
\begin{align}
\label{eq:model_lowlevel_fuzzy_map}    
\fv{c,\ell,k+1}=\Tilde{f}_{\ell}\Big(\tilM,\Mprob{k+1},\mathcal{O}_k\Big) \qquad  \ell = 1, \ldots
, \nF
\end{align}
with $\tilM$ the most recent low-level fuzzy map of size $\nx \times \ny \times \nF$ 
that includes all the fuzzy vectors $\tilvec{c}{k}$ corresponding to all cells $c$ 
with $[\xih_c , \xiv_c]^\top \in \left\{ [1,1]^\top, \ldots, [\nx,\ny]^\top\right\}$, 
$\Mprob{k+1}$ the probability map that has just been updated based on the newly acquired observations, $\mathcal{O}_k$ the set of the most recent observations 
acquired from the environment by all the robots, 
and $\Tilde{f}_{\ell}(\cdot)$ a fuzzy mapping that is used for updating the $\ell^{\text{th}}$ fuzzy 
characteristic that has been associated to each cell in the environment. %
Although \eqref{eq:model_lowlevel_fuzzy_map} is represented in a general way and, thus, contains 
a considerable number of inputs to the fuzzy model $\Tilde{f}_{\ell}(\cdot)$, 
in the implementations for USaR each fuzzy model requires only a few number of the inputs that 
impact each element $\fv{c,\ell,k+1}$.  

The elements of a fuzzy vector corresponding to a low-level fuzzy map may be described by the following characteristics:
\begin{itemize}
    \item 
    \textbf{Binary vs non-binary:} Binary variables are conventionally restricted to values 0 and 1, whereas non-binary variables may adopt intermediate values. A binary map may also be described through fuzzy logic, which extends the Boolean logic \cite{Zadeh:M342}. For instance, the occupancy state of a cell is {“empty” or “occupied”, or its risk state is “safe” or “dangerous”}, with these labels represented by fuzzy sets. 
    \item 
    \textbf{Static vs dynamic:} Static variables depend only on the current measurements, beliefs, and (fuzzy) inputs, whereas dynamic variables (e.g., cell certainty) have a memory and, hence, depend also on their value in the previous iteration.  
\end{itemize}
\bigskip

\begin{remark}
    When used for autonomous mission planning of USaR robots, FLMPC leverages the update equations given by \eqref{eq:model_lowlevel_fuzzy_map} to be embedded in \eqref{eq:cost_func_general} 
    given in the 
    general formulation of FLMPC in \eqref{eq:optimization}.    
\end{remark}
\smallskip

\begin{remark}
    Fuzzy maps in FLMPC are classified as rewards or constraints and are employed to grade 
    candidate trajectories of the robots, as is shown via  \eqref{eq:optimization}. 
    The third category of fuzzy maps in FLMPC are {auxiliary} maps that do not directly affect the cost, but are required to be used within $\Tilde{f}_{\ell}(\cdot)$ to be able to update both reward and constraint fuzzy maps.
\end{remark}

\subsubsection{Grading candidate  trajectories of a robot}
\label{sec:grading_traj}

Each candidate trajectory $\hat{\boldsymbol{x}}_k=\{\boldsymbol{x}_{k+1},\boldsymbol{x}_{k+2},\dots,\boldsymbol{x}_{k+ \np}\}$ for a robot is defined as a 
sequence that contains the coordinates of the cells that are planned at time step $k$ to be visited at time step $\kappa$, with $\kappa \in \{k+1, \ldots, k+\np\}$. 
Based on Assumption~\ref{ass:five}, when following trajectory $ \hat{\boldsymbol{x}}_k$, 
the robot also observes all cells in a neighborhood of radius $\dmax$ of the cells 
{within the trajectory. }
The number of possible trajectories with size $\np$ and starting from the current  
state $\boldsymbol{x}_k$ of the robot is finite. 
We define a set $\Nxhat{k}$, with each element $\elementX{k}$ of it also 
a set that contains one of the possible trajectories for time step $k$, 
as well as all those cells that 
are observed by the robot when traveling through the given trajectory.%

In order to aggregate the satisfaction degrees for all the fuzzy constraints that are 
imposed on the robot in the given environment, as explained in Section~\ref{sec:agg_constraints}, 
\eqref{eq:yager} is used. 
Nevertheless, for aggregating the importance degrees with regard to the fuzzy goals for each candidate trajectory, the following points should be considered: 
\begin{enumerate}[label=\textbf{P\arabic*}]
    \item \label{p1}
    In USaR it is crucial both to maximize the probability of detecting and locating a human through the robot trajectory and to minimize the time of the detection. 
    \item \label{p2}
    In various cases (especially when the environment is not highly dynamic), visiting the same cell on multiple occasions does not increase the overall goal satisfaction. 
    \item \label{p3}
    In USaR, as explained above, not only the cells that embed the robot trajectory play a role in the overall {value of the objective function}, but also do the cells in the neighborhood of these cells (i.e., all cells in each 
    element $\elementX{k}$ of $\Nxhat{k}$).
    \item \label{p4}
    Since the FLMPC problem \eqref{eq:optimization} is in general nonlinear and non-convex, 
    it is difficult, if not impossible, to guarantee the global optimality of the solutions using nonlinear optimization solvers. 
    Thus, we should prevent that the solver ends up in a sub-optimal region of its search region, 
    by generating attraction zones that include high-reward cells and their neighborhoods. 
\end{enumerate}
\smallskip

\noindent
In order to incorporate these points in the aggregation of the rewards for a candidate trajectory, 
we augment the system with additional tuning weights, which belong to $[0,1]$ and per time step tune the degree of importance of visiting the cells in $\elementX{k}\in\Nxhat{k}$ with regard to the fuzzy goals. 
Hence, the tuning weight $\wcom{c,\kc{k}}$ for cell $c \in \elementX{k}$ that is planned at time step $k$ 
to be observed at time step $\kc{k}$, with $\kc{k} \in \{k + 1, \ldots,k+\np \}$, is a function of both the distance $\delta_{c,\kc{k}}$ of the cell from the position of the robot predicted for time step $\kc{k}$ and $\kc{k}$ itself, i.e., 
the number of time steps it takes the robot to visit cell $c$. 
For all $c \in \elementX{k}$ we have:
\begin{subequations}
\label{eq:weight}
\begin{align}
\begin{split}
\label{eq:final_weight}
    \wcom{c,\kc{k}}=
    \end{split}
    \\
    \begin{split}
    \quad
    \left\{
    \begin{array}{ll}
           \max \left\{ \frac{\displaystyle 1}{\displaystyle -\ln \Big(\alpha\left(\delta_{c,\kc{k}}\right)  
    \beta \left(\kc{k}\right)\Big) + 1}, 
    \wcom{c,\kc{k}-1}
    \right\},  
    & \quad \text{if } \alpha\left(\delta_{c,\kc{k}}\right) > 0\\
     \wcom{c,\kc{k} -1},    & 
     \quad \text{if } \alpha\left(\delta_{c,\kc{k}}\right) = 0
    \end{array}
    \right.
\end{split}
\nonumber
\\
\begin{split}
\label{eq:neighborhood}
\alpha \left( \delta_{c,\kc{k}} \right) = \max\left\{0,1- \frac{\displaystyle \delta_{c,\kc{k}}}{\displaystyle \delta^{\max}} \right\}
\end{split}
\\
\begin{split}
\label{eq:time}
    \beta \left(\kc{k}\right)= \gamma^{\kc{k}}
\end{split}
\end{align}
\end{subequations}
where $\gamma \in [0,1]$ is a constant design parameter. 
Note that \eqref{eq:neighborhood} and \eqref{eq:time} incorporate points \ref{p4} and \ref{p1}, respectively. 
The main motivation behind the formulation of \eqref{eq:final_weight} is that the tuning weights should be obtained 
via a monotonic function that assign{s} values in $[0,1]$ to its outputs, associating an equal importance 
to both factors that are formulated via \eqref{eq:neighborhood} and \eqref{eq:time}. 
Since the cell may have been visited more than once, in \eqref{eq:final_weight} the updated value 
of a  tuning weight is 
considered only if it is larger than its previous value, whereas this enables to incorporate point \ref{p2} in the computations.%

Finally, the overall degree $\muG\left(\hat{\boldsymbol{x}}_k\right)$ of satisfaction for the candidate 
trajectory $\hat{\boldsymbol{x}}_k$ for USaR robots $\forall \elementX{k} \in \Nxhat{k}$ is obtained by leveraging \eqref{eq:mean} via: 
{\begin{equation}\label{eq:mean_weight}
   \muG\left(\hat{\boldsymbol{x}}_k,\envir{k}\right) =\left(\frac{1}{\displaystyle\max_{\mathcal{Z} \in \Nxhat{k}}\card{ \mathcal{Z}}}\sum_{c \in  \elementX{k}}\Big(\muG_1\left(c,\envir{k}\right) \star \ldots \star \muG_{\nG}\left(c,\envir{k}\right) \Big)^{\wG+\frac{1}{\wcom{c,\kc{k}}}}\right)^\frac{1}{\wG}
\end{equation}}
Note that the maximization operation in the denominator of \eqref{eq:mean_weight} can be performed offline before the optimization problem of FLMPC is solved. 
This avoids the optimizer to maximize the cost function by reducing the number of the observed cells, which 
is actually an undesirable behavior. 
Finally, by considering the set $\elementX{k}$, instead of the trajectory $\hat{\boldsymbol{x}}_k$, in 
estimation of the overall value of the objective function associated to a trajectory via \eqref{eq:mean_weight}, we effectively incorporate point \ref{p3} in the estimations.%


Once the aggregated importance degree with regard to the goals and the aggregated satisfaction degree with regard to the constraints have been obtained, \eqref{eq:optimization} {is} implemented to determine an 
optimal trajectory for the robot.%

\subsubsection{Coordination of multiple robots}
\label{sec:coop}

In order to explore an unknown environment, utilizing a multi-robot system is common. 
In order to avoid multiple robots visiting the same cells, which hinders the efficiency of the exploration, 
it is essential that these robots are coordinated. In \cite{serialParallel}, three optimization-based coordinating 
methods have been compared for multi-{robot} systems: central control, distributed control with serial communication among the {robots}, and 
distributed control with parallel communication among the {robots}. 
As expected, the central optimization-based approach achieves the best results with regard to optimality, due to its global view of the system. 
The two distributed optimization-based approaches perform very closely, whereas the distributed 
method with serial communication among the {robots} showing a faster convergence rate. 

In this paper, we consider a central system that receives information from all robots regarding their 
current states, observations, and planned trajectories. 
The robots receive updated environmental information from this central system. %
We consider the set $\mathcal{R}$ that includes the indices of all robots. 
Note that in this case the variables that have been defined for a robot in the previous sections are shown 
via an additional subscript index that refers to the robot index. 
Each time a robot has computed their trajectory and tuning weights $\wcom{c,\kc{k,r},r}$ for this trajectory they have planned at time step $\kc{k}$ via \eqref{eq:final_weight} and $\forall r \in \mathcal{R}$, 
these weights are sent to the central system. While cooperating, the robots are given weights computed by other robots and use updated  cooperative weight $\wcomup{c,\kc{k,r},r}$ computed with following equation
{\begin{equation}
    \label{eq:update_tuning_weights}
    \wcomup{c,\kc{k,r},r} = \max \Bigg\{ 0,  \wcom{c,\kc{k,r},r} - \max_{\epsilon \in (\mathcal{R}\setminus\{r\})} \wcom{c,\kc{k,\epsilon},\epsilon} \Bigg\}
\end{equation}}
wherein the tuning weight assigned to a cell by each robot is reduced by the largest tuning weight assigned to that cell by other robots.

This approach bears a resemblance to the previously mentioned distributed controller with serial communication, but a clear distinction emerges. In \cite{serialParallel}, all controllers solve multiple optimization problems, while communicating their decisions through interconnecting variables. The control inputs are only dispatched to robots once convergence had been achieved between all controllers. However, {the exploration problem using a multi-robot system  lacks such flexibility, 
due to the long optimization time despite the  need for fast decisions. 
Instead, once the trajectories are determined, the control inputs steer the robots and 
are sent to the central controller. 
Additionally, each robot needs to know the updated values of the tuning weights for other robots in 
its neighborhood, in order to estimate \eqref{eq:update_tuning_weights}. Thus, these values  are also shared with the robots. }

Nevertheless, this approach has a potential drawback: 
A robot that earlier decides to visit a cell with a high reward may take the chance away from 
another robot whose visit to the same cell could potentially result in a situation that would globally be 
optimal. In other words, this robot may be penalized for attempting to reach the high-reward cell. 
This is the rationale behind the decision to have the central {controller} solve a
general optimization problem, in order to return
a trajectory for each robot. However, this process cannot be executed too frequently,
as it necessitates a significant amount of computational resources, particularly given
the multiplication of decision variables by the number of robots, which presents a
formidable challenge.

\subsubsection{Motion encoding via FLMPC}
\label{sec:encoding}

In our implementation of motion encoding, which {is} explained in Section~\ref{sec:MPC_exploration}, 
two horizons are considered: 
The maximum number $\np$ of the FLMPC decision variables (that is equal to the prediction horizon of FLMPC) and the maximum number $\npath$  of cells that a predicted path consists of.  
Each decision variable includes a sequence of the cells that should be traveled by the robot during the 
corresponding sampling time and the heading values for the robot when traversing each two successive cells in the sequence. 
The number of the cells traveled per sampling time is smaller than or equal to 
a pre-tuned parameter $\ntr$, with $\npath \leq \np \ntr$. 
To avoid over-constraining 
the FLMPC optimization problem, the heading is allowed to continuously vary within interval $[0, 2\pi]$. 
After determining the positions of the robot in cells via the optimization problem, the robot is forced to 
locate itself in the center of the cells.%

The rationale for this design is two-fold: On the one hand, allowing for the maximum number of decision variables, i.e., $\np \ntr$ provides flexibility in generating optimal paths, especially around complex-shaped obstacles. On the other hand, using the same number $\np \ntr$ of decision variables consistently, results in a very high computational load. Accordingly, introducing  $\npath$  
enables to adaptively reduce the number of the decision variables when such an exhaustive exploration 
does not deem necessary (e.g., no complex-shaped obstacles are around the robot).%

Once the FLMPC optimization loop stops and a trajectory is planned for a robot, it is locally assigned an overall fuzzy grade (see Section~\ref{sec:grading_traj}) 
based on the current knowledge of the robot about the environment (i.e., its local maps).  
Then, as it {is} explained in Section~\ref{sec:coop}, these grades are updated according to the plans of other robots that are communicated with the robots via the central system.%

The primary rationale for adopting this methodology rather than conventional priority encoding is that when a robot is positioned in close proximity to the boundaries of the environment, priority encoding is susceptible to selecting a distance in a direction that is contrary to that of the nearest wall. 
This is due to the fact that priority encoding samples the position over the entire environment. 
Consequently, areas in close proximity to the boundaries of the environment or obstacles may not be sampled at all, especially if the environment is of a considerable size. 
This, in USaR, is particularly undesirable since humans may be situated in close proximity to the boundaries, 
but should not remain unattended by the robots.%

\subsection{Bi-level FLMPC based on parent-child architecture}
\label{sec:parent_child_arch}

As robots operate under limited temporal and spatial horizons $\np$ and $\npath$, respectively, in the event of a large and/or highly dynamic environment 
both horizons may need to be extended. Otherwise, upon a relatively short-sighted path planning, 
the robots may overlook targets (e.g., humans) that are located farther away from them. However, 
since the robots should make decisions rapidly, given the environmental variations, 
larger horizons may hinder their applicability.

In \cite{BiLevel}, it {is} proposed to use a bi-level MPC architecture combining short-sighted and long-sighted 
MPC systems in the two levels: On the one hand, the long-sighted MPC makes a plan 
within a larger prediction window {(with a fixed terminal point, i.e., a shrinking -- instead of a rolling -- horizon),} with a larger sampling time (both for decision making and for modeling the dynamics of the controlled system), which makes the computations more efficient. 
On the other hand, the short-sighted MPC operates upon a smaller sampling time to compensate for 
the roughness of the estimations by the long-sighted MPC. 
To afford the computations, the short-sighted MPC deploys a smaller, rolling horizon. 
These two MPC levels interact with each other for decision making via additional constraints 
that are formulated and sent to the short-sighted MPC via the long-sighted MPC, 
to ensure the constraint satisfaction and improved optimality in the longer terms. 
Inspired by this, we propose a bi-level architecture, which we call the parent-child architecture,  
based on the novel concept FLMPC{,} for exploration of large-scale unknown environments via multi-robot systems.%

In the real world, human rescuers are guided by an {``}incident commander{''} who oversees the 
long-term planning and coordination of the rescue team. 
The parent-child FLMPC architecture proposed here allows to deploy a long-sighted (parent) FLMPC system 
to resemble the incident commander for coordination of the robots that operate according to short-sighted 
(child) FLMPC systems. Next, we provide the details about the architecture and functioning of 
the parent-child FLMPC architecture.%

\subsubsection{Mapping for long-term decision making of the parent FLMPC}

As explained above in the introduction of Section~\ref{sec:parent_child_arch}, the main motivation of 
the parent-child architecture is to balance the computational efficiency and 
the future vision of FLMPC. 
Since the parent FLMPC system is long-sighted, to make the computations efficient, 
the map of the parent FLMPC system is made coarsened by clustering multiple cells into one.

All cells in the environment are dynamically clustered, based on fuzzy membership degrees, by the parent FLMPC system.  
Each child FLMPC system, responsible for steering one robot, receives a command from the parent FLMPC system about the next cluster it should explore. 
Per clustering, each cell in the environment is assigned to one or more fuzzy clusters 
with varying degrees of membership. 

These fuzzy clusters are distinguished based on a high-level fuzzy score that is assigned to them 
by the parent FLMPC. This score is indicative of the potential accumulated reward that is associated to the fuzzy cluster, and is estimated by taking a weighted average on the aggregated grades of all 
the cells within that cluster considering their degree of membership to the cluster. 
Therefore, for the score $\score{j,k}$ of the the $j^{\text{th}}$ fuzzy cluster at time step $k$, we have:
\begin{subequations}
\label{eq:scoring_fuzzy_clusters}
\begin{align}
    \begin{split}
        \label{eq:cluster_score}
        \score{j,k} = \left(\frac{\displaystyle \sum_{c=1}^{\nx\ny}\sum_{i = 1}^{\ncluster{k}} (\mucluster{i,k}{c,\envir{k}}\muO_k\left(c,\envir{k}\right))^{w^{\text{cluster}}}}{\displaystyle \sum_{c=1}^{\nx \ny}\sum_{i = 1}^{\ncluster{k}} \mucluster{i,k}{c,\envir{k}}}\right)^{\frac{1}{w^{\text{cluster}}}}, \qquad 
        j \in \left\{1, \ldots, \ncluster{k}\right\}
    \end{split}
    \intertext{where the following equation holds:}
    \nonumber
    \\
    \begin{split}\label{eq:cell_aggregated_score}
        \muO_k\left(c,\envir{k}\right) = \min\left\{\max\big(\muG_1(c,\envir{k}),\dots,\muG_{n^{\text{goal}}}(c,\envir{k})\big),\muC_1(c,\envir{k}),\dots,\muC_{n^{\text{con}}}(c,\envir{k})\right\}
    \end{split}
\end{align}    
\end{subequations}
In \eqref{eq:cluster_score}, $\mucluster{i,k}{c,\envir{k}}$ is the membership degree of cell $c$ 
to the $i^{\text{th}}$ cluster at time step $k$, $\ncluster{k}$ is the total number of the fuzzy clusters determined by the parent FLMPC system at time step $k$,  
and $\nx\ny$ is the total number of the cells in the environment (note that some of these cells may be associated to the $j^{\text{th}}$ fuzzy cluster with a null membership degree).  
The rationale behind \eqref{eq:cell_aggregated_score} 
is the same 
as for \eqref{eq:mean}-\eqref{eq:multiplication}, but instead of a trajectory, here a single cell is considered. 
We have used the index $k$ for the membership degrees to indicate that these degrees may vary in time 
as the global map of the environment is updated by the parent FLMPC system based on the local maps updated by 
{child} FLMPC systems. Also, as before, we use the weight $w^{\text{cluster}}$, which has the same purpose (increasing the importance of {a few high scores} over multiple low scores).%


Each fuzzy cluster is associated with a state at time step $k$ by the parent FLMPC system, 
where this state adopts one of the following $4$ realizations: {“unexplored”, “to be explored”, 
“being explored”, “explored”. }
In case at time step $k$ the fuzzy cluster has never been scanned by any robots before, 
its state is “unexplored”. However, if  
this cluster is within the planning of one or more robots at time step $k$, its state is  updated to “to be explored”. If at time step $k$, one or multiple robots are already exploring the 
fuzzy cluster, its state at this time step is {“being explored”. Finally, if the fuzzy cluster has already 
been explored before time step $k$, the state of this cluster at time step $k$ is “explored”}.%

Whenever the score of a fuzzy cluster determined via \eqref{eq:cluster_score} is less than a specified threshold, designated as $\sexp$, the state of the fuzzy cluster is classified as {“explored”}. 
The score of the fuzzy clusters may increase in the course of the mission, particularly in dynamic environments, 
where the degree of uncertainty associated to a fuzzy cluster tends to increase when it is not subjected to prolonged exploration. 
Consequently, when the score of a fuzzy cluster exceeds a designated threshold value, $\sunexp$, the 
state of that fuzzy cluster is classified as {“unexplored”} once more. 
\smallskip

\begin{remark}
\label{remark:sexp_sunexp}
The value of $\sexp$ should be tuned to be adequately smaller than that of $\sunexp$, 
in order to prevent that the state of fuzzy clusters undergo rapid and arbitrary changes under the influence of random measurements. 
\end{remark}

\subsubsection{High-level long-term planning by the parent FLMPC}\label{sec:parent}

The parent FLMPC system generates a high-level plan for each robot by assigning a path 
to the robot that represents a sequence of fuzzy clusters with state {“}unexplored{”}. 
In order to guarantee that the high-level planning covers the entirety of the environment, 
the parent FLMPC system operates under an additional constraint, i.e., 
each single fuzzy cluster must be included in at least one of the proposed paths for the robots. 
In order to grade the trajectories composed of the fuzzy clusters, 
the parent FLMPC system utilizes \eqref{eq:mean} (instead of \eqref{eq:multiplication} as there are no constraints for the parent FLMPC {system} because they are supposed to be dealt with directly by the child FLMPC {system}, as constraints can be changed rapidly based on { new measurements}), based on \eqref{eq:yager} and 
\eqref{eq:mean_weight}{. However,} instead of the membership degrees for the cells included in 
the trajectories, the scores of the fuzzy clusters involved, as computed by \eqref{eq:scoring_fuzzy_clusters}, are considered. Accordingly, the hyper-parameters $\wG$ may be generalized to  $\wGP$, and may have a different value than $\wG$.%

\textbf{Grading high-level trajectories by parent FLMPC:} 
The sequence of {“}unexplored{”} fuzzy clusters assigned to a robot by the parent FLMPC system 
may include clusters that are not the same as or {are} even far from the current fuzzy cluster that embeds the robot. 
This allows the parent FLMPC system to re-distribute the robots whenever a considerable number of them have been spawned in or around the same fuzzy cluster, 
which impedes the global optimality of the multi-robot system. 
Consequently, the time required to reach the same fuzzy cluster may vary 
for the robots assigned to the same trajectory. 
Moreover, exploring the {“}unexplored{”} fuzzy clusters that are assigned to the robots  
by the parent FLMPC system may require the robots to traverse through intermediate fuzzy clusters 
with a state rather than {“}unexplored{”}. Such clusters, unlike fuzzy clusters of state “unexplored”, 
{do} not contribute as much to the overall reward that is collected by the robot for 
accomplishing the exploration tasks that are assigned to it by the parent FLMPC system 
(see Section~\ref{sec:grading_traj} for details, taking into account that the discussions 
given in that section for the cells should be generalized for the fuzzy clusters when 
discussing the high-level decision making). 
Therefore, these effects should be incorporated in grading the candidate trajectories 
in the high-level decision making. Accordingly, \eqref{eq:weight} for fuzzy cluster indexed $i \in \{1, \ldots, \card{\mathbb{P}_{k}}\}$ that is planned to be 
explored at time step $\kj{k}$ and that belongs to trajectory $\mathbb{P}_{k} = \{[\ph_{1,k},\pv_{1,k}]^\top, \ldots, [\ph_{\card{\mathbb{P}_{k}},k},\pv_{\card{\mathbb{P}_{k}},k}]^\top\}$ determined at time step $k$ by the parent FLMPC system for a robot, is reformulated via \eqref{eq:high_level_tuning_weight}, i.e.:
\begin{subequations}
\label{eq:high_level_tuning_weight}
\begin{align}
\begin{split}
\label{eq:high_level_final_weight}
    \wcomhl{j,\kj{k}}=
           \max \left\{ \frac{\displaystyle 1}{\displaystyle -\ln \left(\beta \left(\kj{k} \right)\right) + 1}, 
    \wcomhl{c,\kj{k} -1}
    \right\},  
\end{split}
\\
\begin{split}
\label{eq:high_level_time}
    \beta \left(\kj{k}\right)= \gamma^{\sum_{i=1}^{j} \norm{[\ph_{i,k},\pv_{i,k}]^\top - [\ph_{i-1,k},\pv_{i-1,k}]^\top}}
\end{split}
\end{align}
\end{subequations}
where $\wcomhl{j,\kj{k}}$ is the tuning weight for cluster $j$ that is planned via the parent FLMPC system at time step $k$ 
to be explored at time step $\kj{k}$ {by the robot}. Note that $\kj{k} \in \left\{k + 1, \ldots,k+ \np \right\}$ and $\card{\mathbb{P}_{k}}$ shows the total 
number of the fuzzy clusters in path $\mathbb{P}_{k}$. Moreover, $[\ph_{i,k},\pv_{i,k}]^\top$ shows the coordinates of the center of the $i^{\text{th}}$ fuzzy cluster in path $\mathbb{P}_{k}$ determined at time step $k$, whereas  $[\ph_{0,k},\pv_{0,k}]^\top$ is the coordinates of the robot that is assigned path $\mathbb{P}_{k}$ by the parent FLMPC system at time step $k$.%


\smallskip 

\begin{figure}
    \centering
    \includegraphics[width=0.5\linewidth]{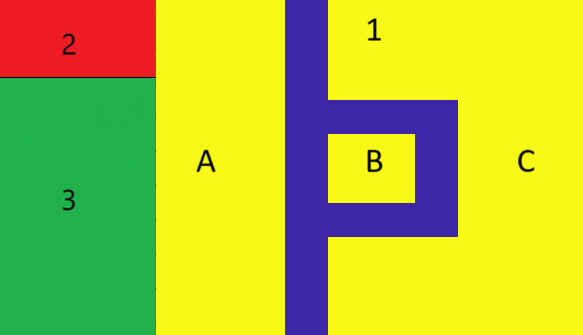}
    \caption{{ Three fuzzy clusters, numbered 1 to 3, are illustrated in different colors: In this case there is}  a wall (shown in dark blue) {that} is identified, leading to the division of {the first} fuzzy cluster into three smaller fuzzy sub-clusters (shown in yellow). {Each sub-cluster is represented by a letter A, B or C.} Given that the sub-cluster {C} is the largest (i.e., it includes more environmental cells), it is considered as the fuzzy cluster {1} that should 
    replace the original fuzzy cluster {1}. 
Since the  fuzzy sub-cluster {B} that is totally encountered by the wall is not connected to any other fuzzy (sub-)clusters, it cannot be included in any trajectories by the high-level FLMPC system. 
The fuzzy sub-cluster {A is} be merged with cluster {3} by the parent FLMPC system. }
    \label{fig:group}
\end{figure}

\textbf{Re-structuring fuzzy clusters with internally inaccessible cells:} 
While exploring a fuzzy cluster, reaching some cells in that cluster may {turn out to} be infeasible from the interior of the cluster. This may be, for instance, due to structures, such as walls, that have not been discerned earlier: 
See Figure~\ref{fig:group}, where the entire rectangular shape shows one fuzzy cluster, in which the cells in the three separate yellow sub-clusters { (A,B,C) }are not accessible through each other, due to the existence of the wall (shown in dark blue). 
Subsequently, such a fuzzy cluster is re-clustered into smaller fuzzy sub-clusters. 
The resulting fuzzy sub-cluster  that includes the largest portion of the cells in the original fuzzy cluster  {(for instance, sub-cluster C in Figure~\ref{fig:group})}
is kept in the original trajectory and replaces the original fuzzy cluster in that trajectory. 
The remaining fuzzy sub-clusters are either merged with other existing fuzzy clusters by the parent FLMPC system, 
or are excluded from the map in case there is no way to reach them based on the current configuration of the robots {(this, for instance, holds for sub-clusters B in Figure~\ref{fig:group})}.%

In order to reallocate the resulting fuzzy clusters that have been discarded from the original trajectory {(for instance, sub-cluster A in Figure~\ref{fig:group})}, 
the parent FLMPC system first assesses whether or not at least one cell of the cluster has previously 
been assigned to any other fuzzy clusters with a non-zero membership degree
{(for instance, in Figure~\ref{fig:group}, there are cells that jointly belongs to fuzzy (sub-)clusters 2 and A, as well to 3 and A). }
Then the mean value for the membership degrees of the cells within the resulting fuzzy cluster to {all other 
such} fuzzy cluster{s} is estimated and at the end, the resulting fuzzy cluster is merged with the fuzzy cluster 
that corresponds to the highest mean value of  membership degrees. 
Note that after being merged, the transferred cells adopt the membership degrees they have with respect to the new fuzzy cluster they are merged into.%

\subsubsection{Low-level short-term planning by the child FLMPC}

Once the high-level plan has been made by the parent FLMPC system, each robot is assigned to a 
trajectory of fuzzy clusters for the purpose of exploration. 
Moreover, the robot receives a local fuzzy map from the parent FLMPC system, updated based on data collected from all robots and using \eqref{eq:model_lowlevel_fuzzy_map}, for the 
region that embeds the given high-level trajectory of the robot.  
The child FLMPC systems implement \eqref{eq:weight}, but to ensure that each robot is able to devote its full attention to its designated fuzzy {cluster}, we transform equation  \eqref{eq:final_weight} into \eqref{eq:final_weight_transformed}, where an additional weight, denoted by $w_{hl}$, is introduced. The value of this weight is dependent upon whether the fuzzy {cluster} is currently in a state {“to be explored” or “being explored”}.

\begin{subequations}
\begin{align}
    \begin{split}
\label{eq:final_weight_transformed}
    \wcomll{c,\kc{k}}=
    \end{split}
    \\
    \begin{split}
    \left\{
    \begin{array}{ll}
           \max \left\{ \frac{\displaystyle 1}{\displaystyle -\ln \Big(\alpha\left(\delta_{c,\kc{k}}\right)  
    \beta \left(\kc{k}\right)\whlC(c)\Big) + 1}, 
    \wcomll{c,\kc{k} -1}
    \right\},  
    & \quad \text{if } \alpha\left(\delta_{c,\kc{k}}\right)\whlC(c) > 0\\
     \wcomll{c,\kc{k} -1},    & 
     \quad \text{if } \alpha\left(\delta_{c,\kc{k}}\right)\whlC(c) = 0
    \end{array}
    \right.
    \end{split}
    \nonumber
    \\
    \intertext{where in case the fuzzy cluster has the state {“}to be explored{”}, we have:}
    \begin{split}
        \label{eq:gauus}
        \whlC(c)=
    \left\{
    \begin{array}{ll}
    \exp\left({\frac{\norm{[\ph_{j,k},\pv_{j,k}]^\top - [\ph_{0,k},\pv_{0,k}]^\top}}{2\sigma^2}}\right), &  \text{if $\norm{[\ph_{j,k},\pv_{j,k}]^\top - [\ph_{0,k},\pv_{0,k}]^\top} < \eta$} \\
        0,  & \text{if $\norm{[\ph_{j,k},\pv_{j,k}]^\top - [\ph_{0,k},\pv_{0,k}]^\top}\geq \eta$ } 
    \end{array}
        \right.
    \end{split}
    \\
    \intertext{Otherwise, in case the fuzzy cluster has the state {“}being explored{”}, we have:}
    \begin{split}
          \whlC(c)=\mucluster{i,k}{c}
    \end{split}
\end{align}
\end{subequations}
where $\wcomll{c,\kc{k}}$ is the tuning weight for cell $c$ that is planned via the child FLMPC system 
at time step $k$ to be explored at time step $\kc{k}$, $\whlC(c)$ is the correction weight that is 
determined for cell $c$ in the course of the child FLMPC planning and via the high-level 
parent FLMPC system, $j$ is the index of the cluster  
that is planned that {for the robot to} explore, 
and $\eta$ is a user-defined threshold.

In the event that the robot is not situated within the designated fuzzy {cluster}, the appropriate course of action is to guide the robot to the cluster. The reward is inversely proportional to the distance of the robot from the center of the fuzzy {cluster}. In this equation, $\norm{[\ph_{j,k},\pv_{j,k}]^\top - [\ph_{0,k},\pv_{0,k}]^\top}$ represents the distance between the robot and the center of the fuzzy {cluster}, while the parameter, designated as $\sigma$ is a variable chosen by the designer and expressed in meters. However, to guarantee that the robot  proceeds in the direction of the selected fuzzy {cluster}, a weight of 0 is applied instead if the robot movements would increase direction to the center of the fuzzy {cluster}. In the event that a robot is already within the fuzzy {cluster}, the membership of the cells is transformed into weights $\whlC(c)$.

\subsubsection{Bi-level architecture}

\begin{figure*}[ht]
\includegraphics[width=\textwidth]{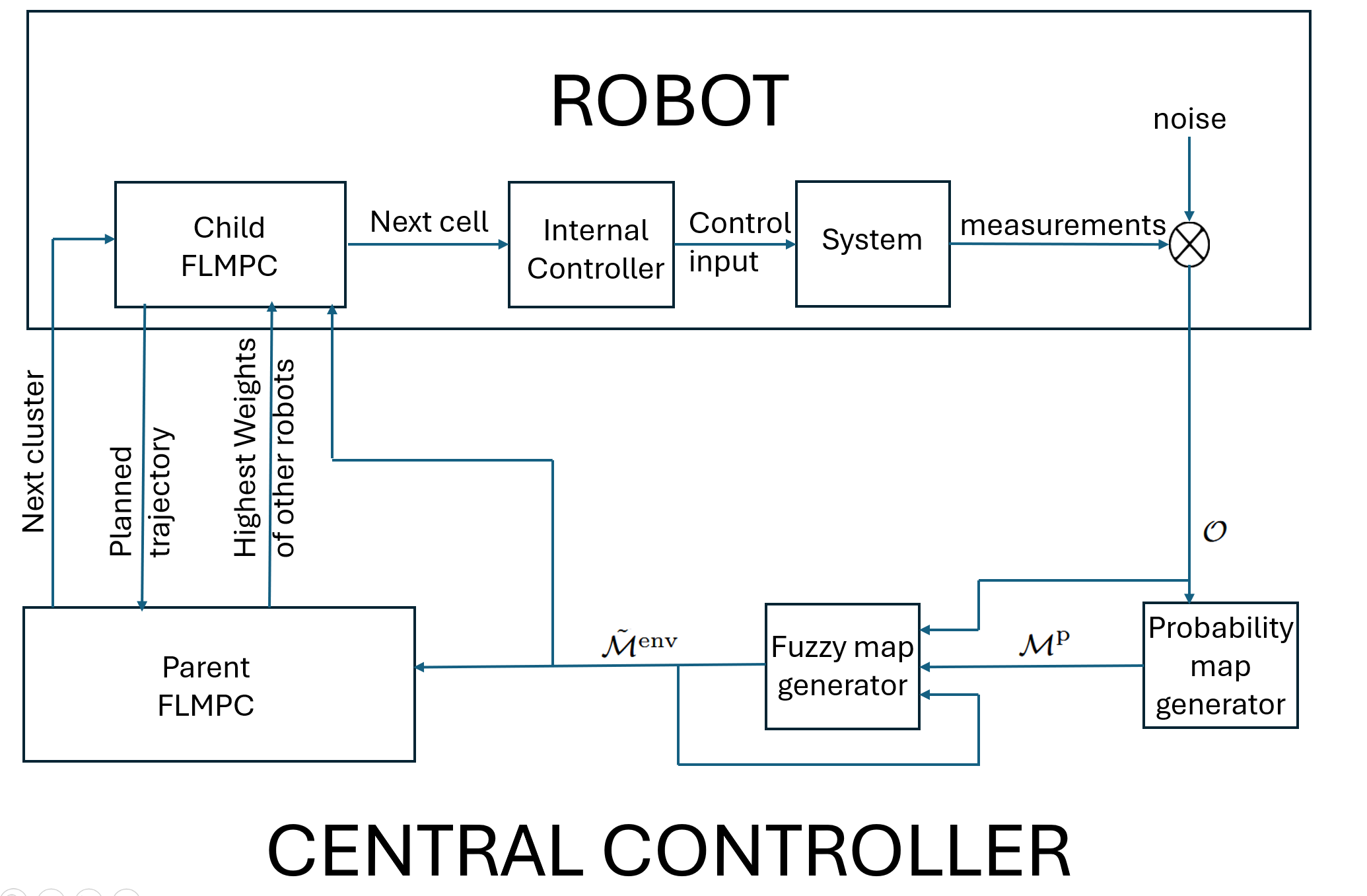}\centering
\caption{The control diagram is presented from the perspective of a single robot.}\label{fig:diagram}
\end{figure*}

To illustrate the operational principles of the system from the perspective of a single robot, we have included a diagram (see Figure \ref{fig:diagram}). Starting from the right-upper corner, we can see that the robot collects measurements that is affected to noise over time. Subsequently, the {central controller} generates a probability map {$\Mprob{k}$} based on the received data. Subsequently, fuzzy maps $\tilM$ are generated based on the measurements, the current fuzzy map, and the probability map. The fuzzy maps are transmitted to the fuzzy {cluster} allocator, which generates the high-level fuzzy scores. These include the computation of both the membership of all cells in each fuzzy cluster (which may change over time) and the score of each fuzzy {cluster}. The long-term FLMPC then sends the next fuzzy {cluster} to be explored. The {cluster} is calculated at the beginning of the mission and again each time the currently explored fuzzy {cluster} is finished.

The short-term FLMPC is connected to a robot and requires knowledge of the current {cluster} to which it is allocated, the trajectories of other robots, and the fuzzy maps. Based on the aforementioned knowledge, a planned trajectory of cells to be visited is computed. Subsequently, the planned trajectory is transmitted to the {central controller}, thereby enabling other {robots} to become acquainted with the local {robot}'s plan and circumventing a scenario in which two robots attempt to rescue the same individual. The short-term FLMPC then transmits the reference $r$ to the short-term controller, which is presumed to be equipped with the robot. Ultimately, the lowest-level controller oversees direct robot control through inputs $u$.  

{Algorithm~\ref{alg:general} provides} a comprehensive understanding of the operational principles of the central controller. The specific instructions for implementing each line of the pseudo-code {given in Algorithm~\ref{alg:general} will be detailed in Section~\ref{sec:implmentation}, Case Study.}

\begin{algorithm} 
\caption{This algorithm shows how the parent FLMPC {system} works for search and rescue robotics from the point-of-view of the {central controller}.}\label{alg:general}
\begin{algorithmic}
\Require An unknown environment, robots are spawned in known cells
\State Generate initial probability map
\State Generate initial fuzzy map{s}
\State Generate initial fuzzy clusters estimating the scores via \eqref{eq:cluster_score}
\While{Mission is not finished}
\State Generate new high-level plan via parent FLMPC 
\State Allocate new trajectories of fuzzy clusters to robots {that are} not assigned to a trajectory
\While{{scores for all fuzzy clusters are above a given threshold}}
\For{each robot in $\mathcal{R}$}
        \State {Send fuzzy maps to the robot}
        \State Compute all{ weights used in \eqref{eq:update_tuning_weights}}
        \State Send {the weights to the robot}
        \State Get planned trajectory {from the robot}
        \State Get measurements { from the robot}
\EndFor
\State Update probability map
\State Update fuzzy maps
\If{a new{ “wall” }is found}
\If{it is required to divide a fuzzy cluster into smaller fuzzy clusters}
    \State Allocate or delete resulting fuzzy clusters

\EndIf
\EndIf
\State Update scores of fuzzy clusters
\EndWhile
\EndWhile
\end{algorithmic}
\end{algorithm}

\section{Case study}\label{sec:implmentation}

This section presents two case studies: 
The first one compares an MPC-based controller with a stochastic cost function against one using a fuzzy goal, evaluating their performance based on the time required to detect all humans in the environment and the computation time. Both controllers utilize the same optimization solvers, encoding methods, probability map updates, and parameters. 
The second case study expands the analysis by comparing the performance of 
single-level and bi-level parent-child FLMPC architectures in a larger environment.%

After evaluating various general-purpose nonlinear optimization solvers 
from the global optimization toolbox of MATLAB, the particle swarm algorithm  
\cite{MatlabPS,ParticleSwarm} proved to be the most effective for this problem, 
balancing both the realized values for the cost function and the computation time.%

The code used to generate the data and results (including figures) is available in \cite{codeFLMPC}. 
To run all simulations efficiently within a reasonable time frame, we used the Delft Blue supercomputer at Delft University of Technology \cite{delftBlue}, which provides $64$~CPU cores per node with a high memory throughput of $250$ GB RAM.%

\subsection{Case study 1: Stochastic cost functions versus fuzzy goals}

\begin{figure}[ht]
\center
\includegraphics[width=0.5\textwidth]{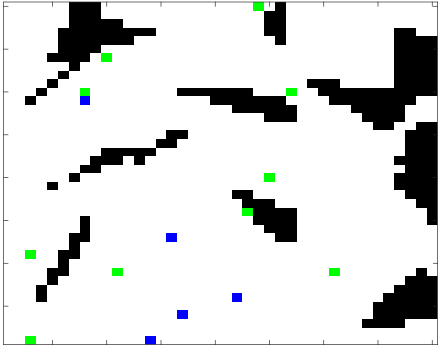}
\caption{A randomly generated environment for the first case study: In this environment, white is used to represent empty spaces, black is used to represent the obstacles, green is used to represent humans, and blue is used to represent the robots. }\label{fig:smal_env}
\end{figure}

To evaluate the performance of FLMPC, this section presents an illustrative case study 
focused on a relatively straightforward problem, i.e., locating humans in an unknown environment using a multi-robot system in the shortest possible time. Each environment is randomly generated, and, except for its size, is unknown. 
The experiment is repeated across $100$ randomly generated environments, comparing two controllers: 
MPC with a stochastic cost function based on prior work \cite{priorityEncoding} and a single-level FLMPC. 
While randomness in the initial placement of humans is inherent in the search process, this variability 
is mitigated through multiple simulation runs. 
Figure~\ref{fig:smal_env} illustrates an example of a randomly generated environment.%

Both controllers use the same parameters: $\np$=5, $\ntr$=14 and $\npath$=20. 
Moreover, to maintain a balanced trade-off between computational efficiency and coordination of the robots, 
the central controller is executed every 5 time steps.%

\subsubsection{Problem setup}

The environment is composed of $40 \times 40$ cells, i.e. $n^{\text{h}}=n^{\text{v}}=40$.  Each cell in the environment is described by a state from the ordered set $\mathcal{S} = $\{empty, human, obstacle\} (i.e., $\rho_1$ = empty, $\rho_2$ = human, $\rho_3$ = obstacle). No prior information about the real states of the environment is available. The mission commences with the random placement of $n^{\text{rob}}$ robots, with $2 \leq n^{\text{rob}} \leq 10$.%

Per step, each robot may traverse one cell in any direction, including diagonal. The 
movement of the robots is unconstrained. Upon reaching a cell containing a human, the human is deemed rescued, and the cell is rendered empty afterwards. If a robot attempts to access a cell with an obstacle, the movement is canceled. 
Subsequent to the completion of each movement, the robots obtain data regarding cells in a neighborhood of maximum $\delta^{\max}=6$ cells. 
Measurement accuracy decreases as the distance between the robot and a cell increases. To apply  \eqref{eq:sensor_model} to cell $c$ where the robot is located at time step $k$, the following relationship is used model the perception accuracy: 
\begin{equation}
\label{eq:sensor_model_cs}
p\left(o_{c,k}|\rho_i\right)=
\left\{
\begin{array}{ll}
0.02 & \qquad o_{c,k}\neq \rho_i 
\\  
0.96     &  \qquad o_{c,k} = \rho_i  
\end{array}
\qquad 
    \forall \left(o_{c,k},\rho_i \right) \in \mathcal{S}\times \mathcal{S}
\right.
\end{equation}
The perception accuracy for neighboring cells within the perception field of the robot depends on their distance from the cell where the robot is located. 
The larger the distance, the higher the uncertainty of the observations. 
To introduce the impact of the distance, detectability $d_{c^{\prime},c,k}$ of cell $c^\prime$, which is $\delta_{c^\prime,\kappa_{c^\prime,k}}$ cells away from cell $c$ where the robot is located and is planned to be observed by the robot at time step $\kappa_{c^\prime,k}$, at time step $k$ is formulated by:
\begin{equation}\label{eq:range_penalty}
    d_{c^{\prime},c,k}
    = \max\left\{1 - \left(\frac{\delta_{c^\prime,\kappa_{c^\prime,k}}}{\delta^{\max}}\right)^2,0\right\}
\end{equation}
Cell $c^\prime$ with a non-zero detectability is perceived by the robot, where 
the perception accuracy $\forall \left(o_{c^\prime,k},\rho_i \right) \in \mathcal{S}\times \mathcal{S}$ is determined via:
\begin{equation}
\label{eq:penalized_sensor}
    p\left(o_{c^{\prime},k}|\rho_i\right)=p\left(o_{c,k}|\rho_i\right)d_{c^{\prime},c,k}+0.25 \left(1-d_{c^{\prime},c,k}\right) 
\end{equation}
where \eqref{eq:penalized_sensor} holds only for cells with  a positive  detectability value.
Obviously when $c^\prime$ and $c$ refer to the same cell, the detectability of the cell according to 
\eqref{eq:range_penalty} is $1$ and thus, \eqref{eq:penalized_sensor} boils down to \eqref{eq:sensor_model_cs}.%

For simplicity, we assume a static environment, where { humans and obstacles} do not move. Simulations continue until all humans are rescued or a predefined time limit is reached, calculated by $500/n^{\text{rob}}$ time steps based on the number of deployed {robots. 
Extensive simulations confirm that this termination criterion prevents outliers, 
such as a robot failing to receive reliable measurements,} from skewing the results. 
As robots lack initial {knowledge about the environment}, their initial belief vector for each cell is set to $[0.34,0.33,0.33]{^\top}$.

\subsubsection{MPC with a stochastic cost function}

The goal {is} to replicate the algorithm from \cite{priorityEncoding}, but  
modifications are required due to {}differences with our case study. The original cost function {in \cite{priorityEncoding} is a weighted summation} of four terms (search effectiveness, uncertainty, detection response, and motion cost). 
For this case study, the first two terms {are} retained, but the detection response is removed due to a higher likelihood of false human detections. 
Without this modification, {robots may move toward falsely reported human locations. To address this issue, a new term is introduced to reward robots for reaching cells where the likelihood of presence of a human is close to $100\%$}.  
Moreover, in the absence of maneuvering penalties, the motion cost term is unnecessary. Unlike the obstacle-free environment {in \cite{priorityEncoding}}, 
an additional constraint is required in this case study, 
i.e., a penalty function that increases the cost when a robot enters a cell with a high probability of containing an obstacle.

\subsubsection{FLMPC}

Designing FLMPC involves three key steps, i.e., defining the fuzzy variables, 
modeling them using membership functions, and setting hyper-parameters 
to regulate the behavior of the controller. 
For this case study, $5$ fuzzy variables, namely ``passability'', human detection reward'', 
``exploration reward'', ``uncertainty'', and ``measurement consistency'' have been considered 
(see Table~\ref{tab:variables}).

\begin{table*}\caption{{List of all fuzzy variables used by FLMPC}}
\label{tab:variables}
\begin{center}
\begin{tabular}{lccc|c}
                        & Reward & Constraint  & Auxiliary  & Dynamic  \\
                        \hline
Passability             &          & \checkmark                 & &       \\
\hline
Human detection reward            & \checkmark                    &   & &           \\
\hline
Exploration reward      & \checkmark        &                     & & \\
\hline
Uncertainty             &          &              & \checkmark       &  \checkmark  \\
\hline
Measurement consistency &          &              &                 \checkmark & \\
\hline
\end{tabular}
\end{center}
\end{table*}

\begin{figure*}[ht]
\includegraphics[width=\textwidth]{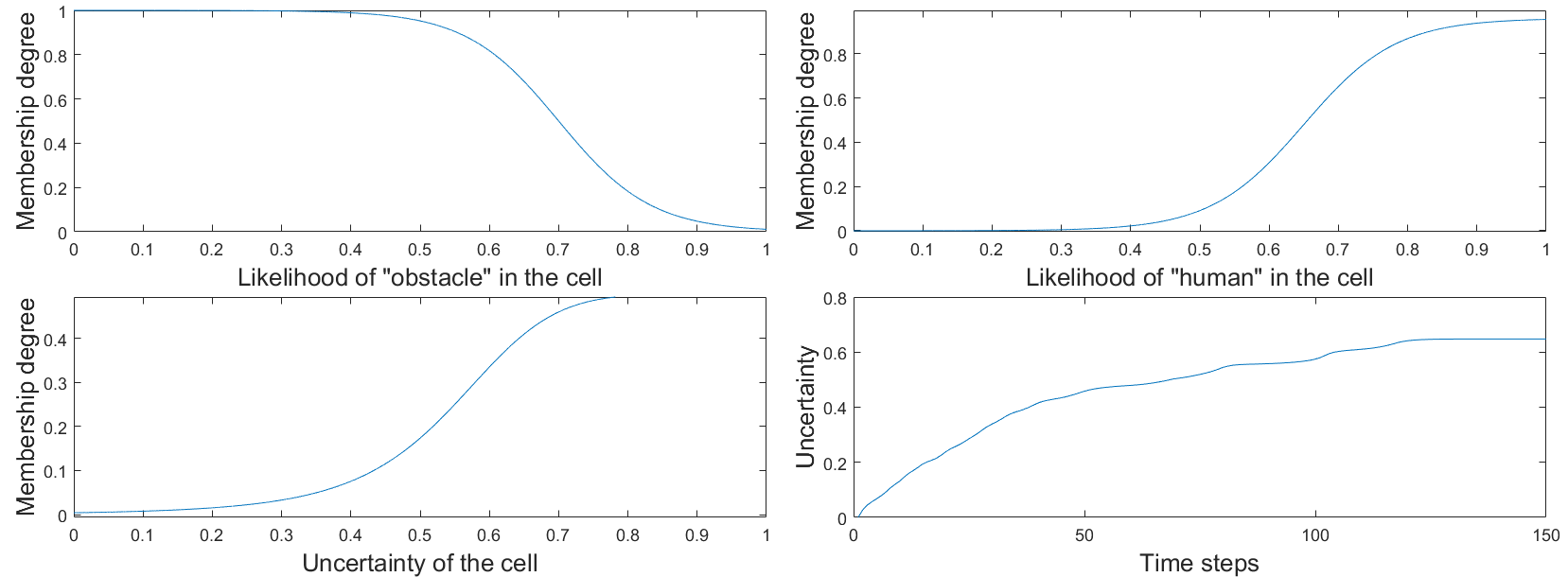}
\caption{{ Top left: Membership function for the passability of a cell, given the likelihood that the state of the cell is ``obstacle''. Top right: Membership function for human detection reward for a cell, given the likelihood that the state of the cell is ``human''. Bottom left: Membership function for exploration reward for a cell, given the uncertainty of the cell within $[0,1]$. Bottom right: Evolution of the uncertainty degree for a cell; uncertainty degree raises over time as long as the cell is not visited/observed.}}\label{fig:pass}
\end{figure*}

The rewards and constraints of the environment are identified based on the goals and setup of the case study. 
Locating humans is the primary objective in search and rescue. 
Thus, ``human detection reward'' is one of the fuzzy variables that is chosen as a reward. 
Moreover, incorporating additional goals improves 
the behavior of robots, as evidenced by other approaches that are based on 
multiple goals (see, e.g., \cite{priorityEncoding,antColony}). 
We, thus, additionally introduce the fuzzy variable ``exploration reward'' for 
exploring the unknown environment. For the constraints, 
as the robot can only interact with passable cells, the fuzzy variable ``passability'' has been introduced.  
For the fuzzy variables ``passability'', ``human detection reward'', and 
``exploration reward'', a static model with a single input suffices.  
For instance, for ``passability'' the membership function that is shown in Figure~\ref{fig:pass} (top left) 
is used, where the input is the likelihood that the state of a cell is ``obstacle''. 
``Human detection reward'' depends on the probability of successfully locating a person, 
i.e., the robot should distinguish between actual {victims and observations that are wrongly reported as humans}.
The membership function used for fuzzy variable ``human detection reward'' is illustrated in Figure~\ref{fig:pass} (top right).  
Without prior knowledge about the location of humans, the robot should be 
encouraged to explore cells with higher uncertainty. Thus, the membership function 
for ``exploration reward'' receives the uncertainty of a cell as input (see Figure~\ref{fig:pass}, bottom left).%

\begin{remark}
    {The maximum degree for the ``human detection reward'', as shown in Figure~\ref{fig:pass} (top right), is intentionally set below $1$. 
    This, based on \eqref{eq:mean_weight}, ensures that tuning weights can still   
    influence the decisions of robots, an effect that disappears when the membership degrees in \eqref{eq:mean_weight} reach $1$. The impact of the tuning weights, as it was explained in Section~\ref{sec:grading_traj}, 
    is crucial in penalizing robots for delaying a rescue.}
\end{remark}

\begin{figure*}[ht]
\includegraphics[width=\textwidth]{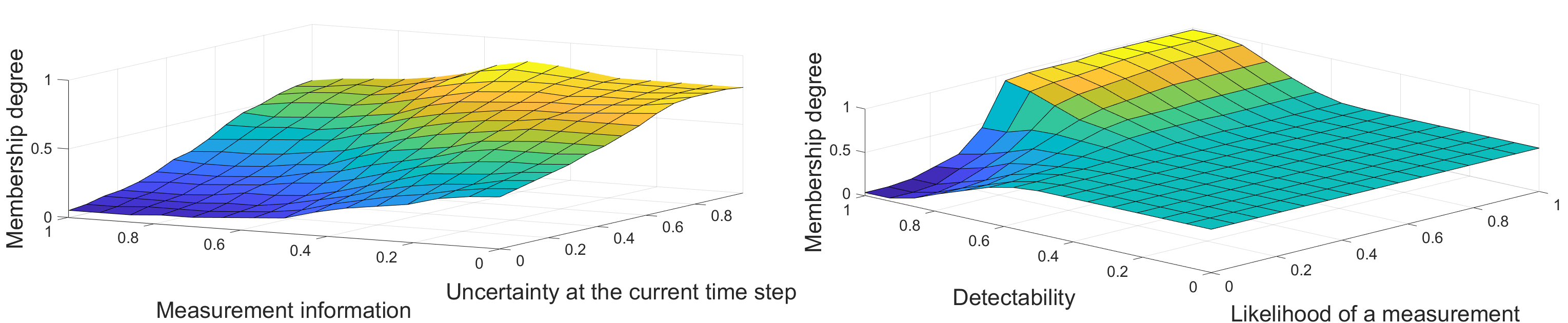}
\caption{Left: Membership function for ``uncertainty'' when the cell is observed, given the most recent value for the uncertainty degree and for the  measurement consistency degree. Right: Membership function for ``measurement consistency'', given the detectability of the cell and the likelihood for a measurement.}\label{fig:visited}
\end{figure*}

For the fuzzy variable ``uncertainty'' a multi-dimensional membership function is used.  
{Although some papers \cite{fuzzyMPCsearch,priorityEncoding,searchUAVlearning}} model the uncertainty as a function of the cell detectability only, this approach faces limitations. { Notably, it does not account for conflicting measurements. 
In fact, if new measurements conflict with existing cell knowledge, the inference about the 
uncertainty should differ from cases where observations align. 
This prevents robots from prematurely considering a cell as explored, when recent 
measurements do not provide a clear conclusion.} 
To account for the consistency of the measurements obtained for a cell, 
we deploy a dynamic fuzzy model for uncertainty with two inputs: most recent certainty degree 
and current measurement consistency degree (see the left plot in Figure~\ref{fig:visited}). 
``Measurement consistency'' is an auxiliary fuzzy variable that reflects whether the new measurement confirms or contradicts the existing knowledge (see the right plot in Figure~\ref{fig:visited}). 
Its membership function receives two inputs: detectability of the cell and the 
likelihood for a measurement.   
When the measurement consistency is below $0.5$, the measurement is considered as not aligned with prior knowledge. This appears as small to no impact for the new measurements on the updated 
uncertainty degree (see the left plot in Figure~\ref{fig:visited}).

Initially, the membership degrees of uncertainty for all the cells are set to $1$, 
which gradually decreases by gathering measurements for the cells. 
Figure~\ref{fig:pass} (bottom right) represents the evolution of the uncertainty degree of a cell 
when no measurements are received for the cell. The uncertainty degrees smaller than $0.648$ increase slowly until reaching the maximum value of $0.648$. 

 In the implementation of the case study, we use $\gamma=0.965$, $\delta^{\max}=5$,  $w^{\text{goal}}=20$, $w^{\text{con}}=5$, $w^{\text{agg}}=1$, and $\np = 5$ in \eqref{eq:yager},\eqref{eq:multiplication},\eqref{eq:weight},\eqref{eq:mean_weight}.%

\subsubsection{Results for case study~1}

We compare the results of {the MPC controller with a stochastic cost 
function and the FLMPC} controller, starting with computational complexity. 
Although both methods use the same number of particles and iterations 
in the optimization solver, the time needed to find an optimal solution differs significantly. On average, MPC with a stochastic cost function took more than 
$60$ times longer than FLMPC. 
This difference arises because grading a single trajectory in MPC with a stochastic cost function demands substantial computation. 
Updating the probability and certainty maps is particularly time 
consuming compared to solving optimization problem \eqref{eq:optimization}. 

We compare the performance of the controllers in terms of both speed and consistency in rescuing the humans. For a fair comparison, simulations for $100$ environmental conditions were performed  
to account for the randomness related to initializing the positions and decisions of the robots (when 
there is not enough information yet to make decisions via the controller).  
Comparing the mean completion times directly for the environmental setups is problematic, since the time required to finish a simulation is influenced by the generated environment and initial conditions. What seems like a short time in one scenario may be long in another, making a simple mean duration comparison ineffective. 
Instead, we assess how often each controller rescues people faster than the other controller and the time difference between them. 
In a simulation, we define a winner as the controller that first reaches a chosen milestone (e.g., saving $6$ out of $10$ humans). The advantage is the absolute time difference between  both controllers in reaching the milestone. If one controller reaches the milestone before the simulation ends while the other one does not, the advantage is considered infinite.

\begin{figure}[ht]
\includegraphics[width=1\textwidth]{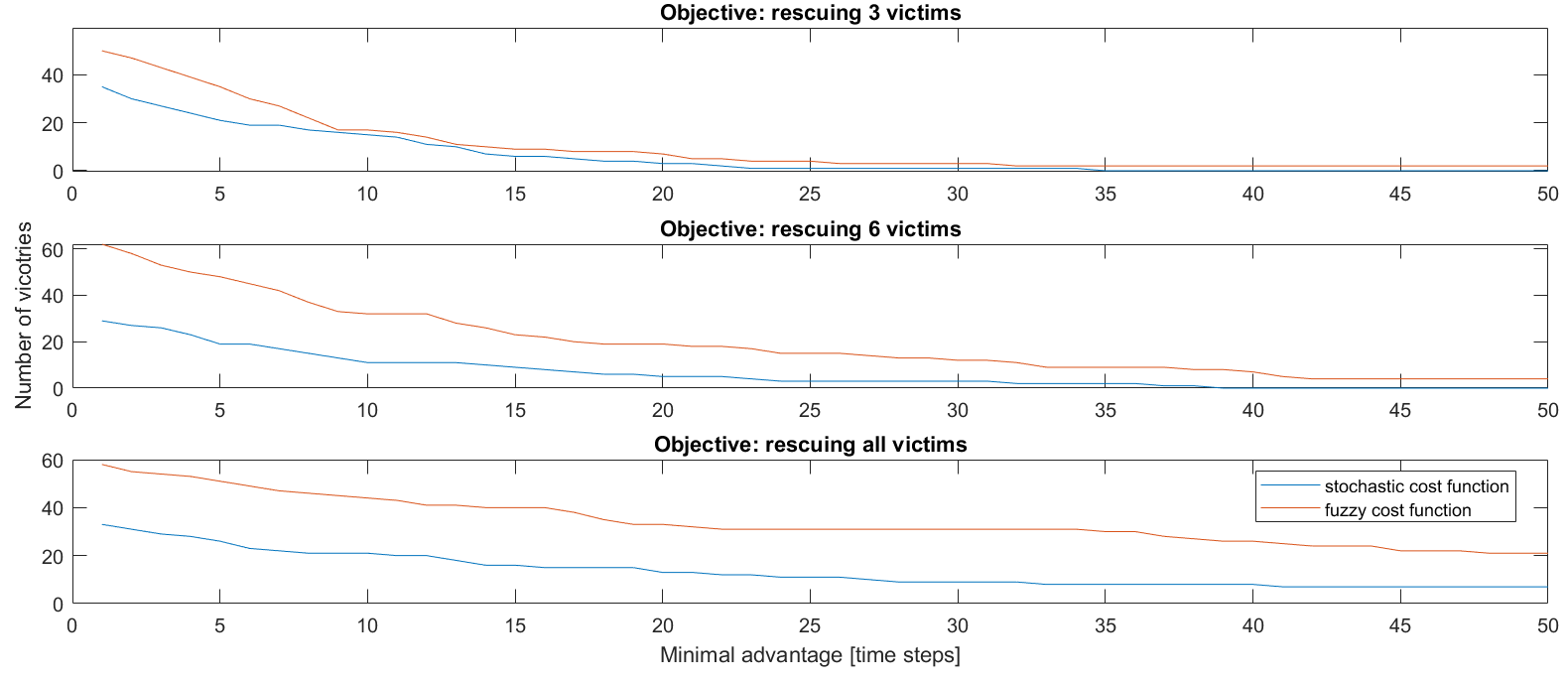}
\caption{{Number of times that each controller is the winner for a specific milestone, shown above each figure, and its advantage. For instance, in $20$ simulations FLMPC is wins in saving $7$ victims with an advantage of at least $24$ time steps.}}\label{fig:wins}
\end{figure}

Figure~\ref{fig:wins} illustrates the win frequency of each controller and 
the advantage gained. It also presents results for  two intermediate goals, next to the goal of rescuing all victims. 
The plots show that FLMPC outperforms MPC with a stochastic cost function in $6$ out of 
$10$ cases. Moreover, FLMPC achieves \textbf{four times more victories} with a significant advantage of at least $35$ time steps compared to MPC with a stochastic cost function. This indicates that FLMPC not only wins more frequently, but is also 
much more efficient at locating and rescuing victims in significantly less time.%

\subsection{Case study 2: Comparison between single-level {child} FLMPC and bi-level {parent-child} FLMPC}

To highlight the importance of a bi-level {parent-child} approach, this section examines the same problem as in the first case study but in a larger environment of $40000$ cells, with $200$~cells in each direction and a simulation limit of $3000$ time steps. 
The environment was randomly generated, while the number of robots remained $3$. A bi-level parent-child FLMPC system was compared to a single-level FLMPC system that replicates the child FLMPC level. 

Per time step and for each robot, the child FLMPC optimization problem considers the cells within an area of $51$~cells $\times$ $51$~cells, centered on the robot. This means that each robot makes decisions based on less than $7\%$ of the cells. 
This is expected to impact the global performance of the single-level FLMPC system. 
The only distinction between the single-level child FLMPC system in this case study and the FLMPC system from the first case study lies in how the fuzzy uncertainty map is updated. 
If a cell is not observed, its uncertainty degree increases $100$ times more slowly than before to prevent a rapid loss of the certainty in early stages of exploration. 

\begin{figure}[ht]
\includegraphics[width=0.6\textwidth]{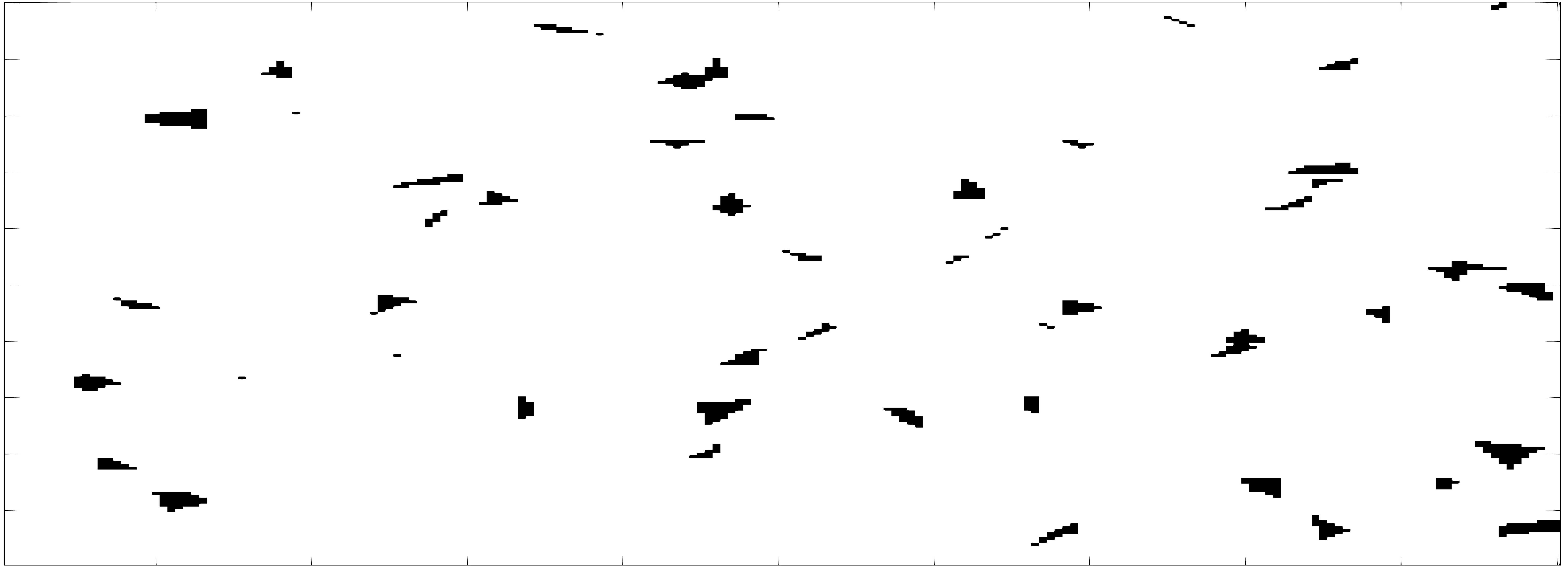}\centering
\caption{{A randomly generated environment for the second case study: In this environment, white is used to represent empty spaces, black is used to represent the obstacles, green is used to represent humans, and blue is used to represent the robots. }}\label{fig:large_env}
\end{figure}

\subsubsection{Bi-level FLMPC}

To implement the bi-level parent-child FLMPC architecture, the environment is divided into $25$ square-shaped 
sub-areas including $40$~cells per side (i.e., each sub-area is of the same size as the entire environment 
in the first case study). Then fuzzy clusters are initialized by including one of these sub-areas, where 
the membership degree $\mucluster{i,0}{c, \envir{0}}$ of each cell $c$ of this sub-area in the $i^{\text{th}}$ cluster  is initialized to $1$. 
To enhance flexibility, allowing robots to exit their designated fuzzy cluster when beneficial, 
$3$ cells in the environment from the edge of the square-shaped sub-area are also included in the fuzzy cluster, 
with a membership degree given by: 
\begin{equation}\label{eq:group_membership}{
    \mucluster{i,0}{c,\envir{0}} = \left(1-\frac{\delta^{\text{cluster}}_{c,i,0}}{4}\right)^2}
\end{equation}
where $\delta^{\text{cluster}}_{c,i,0}$ is the distance of the cell from the edge of fuzzy cluster $i$ at time step $0$ (i.e., $\delta^{\text{cluster}}_{c,i,0} \in \{1, 2, 3\}$). 

For long-term planning via the parent FLMPC system (see Section~\ref{sec:parent}), 
the state of each cluster should be determined. We use $\sexp=0.22$ and $\sunexp=0.4$ 
(see Remark~\ref{remark:sexp_sunexp}) and 
initialize the score of the clusters slightly below $0.5$, since the cells are initially 
unexplored and there is no information about victims. 
If the passability of a cell is below $0.1$, the cell is treated as an obstacle 
within the fuzzy cluster. 
If there is a row of such cells, the fuzzy cluster is divided into sub-clusters (see Figure~\ref{fig:group}). Moreover, we use  $w^{\text{cluster}}=5$ in \eqref{eq:cluster_score}, 
$\wGP=1$ as is explained in Section~\ref{sec:parent}, $\gamma=0.98$ in  \eqref{eq:high_level_time}, and $\sigma=60$ in \eqref{eq:gauus}. 

General solvers struggle with solving the parent FLMPC optimization problem, mainly due to the constraint that each fuzzy cluster must be assigned to exactly one robot. 
The problem closely resembles the vehicle routing problem \cite{VRP} 
or the multiple traveling salesman problem \cite{mTSP}, but unlike traditional formulations of these problems that focus on minimizing time or distance, 
the parent FLMPC optimization problem involves a multi-objective cost function. 
To efficiently solve this problem, we leveraged a genetic algorithm, which offers fast convergence and extensibility, inspired by the one in \cite{mTSPga}. 
This algorithm introduces customized mutation and crossover operators for the multiple traveling salesman problem. 
We have adapted an existing MATLAB implementation given in \cite{mTSPcode} for 
the parent FLMPC multi-objective optimization problem that ensures to optimize 
the assignment of robots with respect to both exploration efficiency and rescue success rates.%

\subsubsection{Results for case study~2}

The results for $4$ representative simulations of large-scale search and rescue missions (called mission~1, \ldots, mission~4) are summarized in Table~\ref{tab:times}, giving the completion time for each mission 
under both the single-level child FLMPC system and the bi-level parent-child FLMPC system. 
For all these cases, both controllers finished the mission (i.e., located all the victims) before the given time budget 
(i.e., $3000$ time steps).%

The bi-level parent-child FLMPC method demonstrated superior performance in $2$ out of the 
$4$ simulations (i.e., mission~1 and mission~3). A significant observation was that the bi-level parent-child FLMPC 
system consistently outperformed the single-level FLMPC system whenever the latter 
struggled to locate the last victim. 
This suggests that the coordination offered to the bi-level parent-child architecture 
by the parent level plays a crucial role in eliminating inefficiencies in the later stages of the mission, leading to a more balanced 
and globally optimized exploration strategy compared to the single-level child FLMPC system. 
In fact, the single-level FLMPC system, due to its inability to 
account for information across the entire environment, tends to adopt a fast and greedy exploration approach. 
In contrast, the bi-level parent-child FLMPC system considers the entire environment, leading to 
a more deliberate and cautious exploration strategy, ensuring that no victims are overlooked. 
The single-level FLMPC system may locate humans faster in some missions, but this is not guaranteed and mainly depends on how well exploration aligns with the environmental setup.%


\begin{table*}[]\caption{Comparison of the completion time of each mission (the values are given in time steps).}
\begin{tabularx}{\linewidth}{X|llll}
                            & Mission 1         & Mission 2         & Mission 3         & Mission 4              \\
                            \hline
Single-level FLMPC & 2869 & 2059 & 2688 & 2222   \\
Parent-child FLMPC    & 2016 & 2435 & 2447 & 2664
\\
\hline
Improvement due to adding the parent level &  $29.73 \%$ & $-18.26\%$ & $8.97\%$ & $-19.89\%$ 
\end{tabularx}\label{tab:times}
\end{table*}


To illustrate this problem, we analyze two cases from Table~\ref{tab:times}: Mission 1 and Mission 4, where the Bi-level parent-child FLMPC system, respectively, wins with highest advantage and loses with highest disadvantage.

\begin{figure}[ht]
\includegraphics[width=0.8\textwidth]{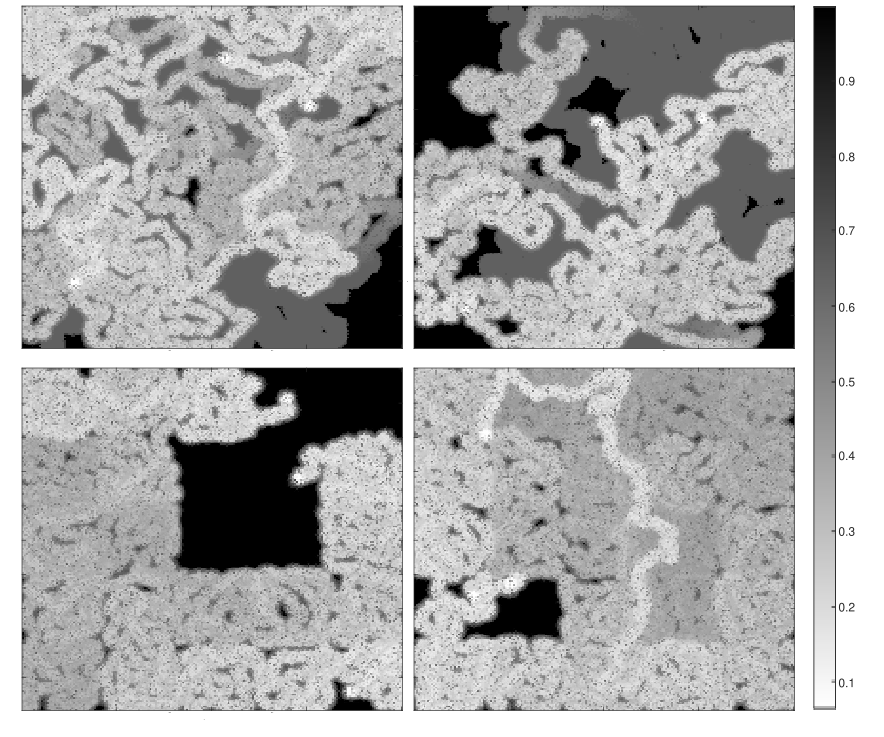}\centering
\caption{{End-of-mission uncertainty map for mission~1 under the single-level FLMPC system (top left),  mission~4 under the single-level FLMPC system (top right), mission~1 under the bi-level parent-child FLMPC system (bottom left), and mission~4 under the bi-level parent-child  FLMPC system (bottom right). The brightness of the pixels represents the uncertainty.}} \label{fig:single_env1}
\end{figure}

In mission 1, as shown in Figure~\ref{fig:single_env1} (top left), the single-level FLMPC system 
leads to chaotic search behavior due to its greedy exploration. 
Despite significant time passing, large portions of the environment remain unexplored. 
Furthermore, by the end of the mission, only one robot continues searching, while 
the other two robots are trapped within areas that are already explored. 
Consequently, when humans are located in the remaining unexplored area, it takes 
substantial extra time to rescue them. 
Conversely, the bi-level parent-child FLMPC system follows a more structured and 
coordinated approach (see the bottom left plot in 
Figure~\ref{fig:single_env1}). This organized search pattern ensures that the locations of 
victims have less impact on the overall search time.%

In mission~4,   
despite the unfortunate circumstances, the the bi-level parent-child FLMPC system ensures that most of the environment 
is explored (see the uncertainty map just before discovering the last victim 
in the bottom right plot of Figure~\ref{fig:single_env1}). This significantly reduces the impact of random exploration on the worst-case mission time. 
Moreover, within the remaining unexplored region, two 
robots are actively searching, enduring that even in the worst-case mission for the 
bi-level parent-child FLMPC system, 
the additional time required to complete the mission is only marginally higher than the actual mission time. 
In contrast, while the single-level FLMPC system completes the mission faster (see the top right plot 
in Figure~\ref{fig:single_env1}), it remains highly sensitive to locations of the victims. 
In case victims are instead located in the  large portions of the environment that remain unexplored, 
the mission takes considerably longer. 
This highlights a key drawback for the single-level FLMPC system, i.e., its worst-case mission time  
is significantly higher than the bi-level parent-child FLMPC system, 
making its performance less reliable in unpredictable scenarios.%

\begin{remark}
\label{rem:limitation_bilevel_FLMPC}
In some additional simulations, the bi-level parent-child FLMPC system 
failed to complete the mission within the time budget. This occurred because certain fuzzy clusters 
were prematurely marked as ``explored'', as high-detectability cells without victims increased certainty, 
while low-detectability cells containing victims remained unexplored. 
\end{remark}



\section{Conclusion{s} and future work}\label{sec:conclusion}

This paper introduces a novel control concept based on fuzzy optimization, 
called fuzzy-logic-based model predictive control (FLMPC), 
and leveraging FLMPC, presents a multi-robot autonomous exploration approach 
for search and rescue in unknown environments.  
While conventional model predictive control (MPC) methods commonly employ stochastic cost functions to handle 
uncertainty in unknown environments, we propose dynamic fuzzy maps that systematically 
process and structure high-level environmental information. 
Unlike stochastic cost functions, which require extensive sampling and recalculations, 
by providing a structured way for capturing the uncertainties via fuzzy maps 
and by incorporating these fuzzy maps into the objective function of FLMPC, we  
enhance the computational efficiency of decision making and the 
success rate of multi/single-robot search and rescue missions, 
while maintaining robust adaptability to dynamic environments. 
As demonstrated by the results of the first case study, replacing the stochastic cost function 
not only improves the performance by reducing the mission completion time, 
but also significantly decreases the computational complexity (for up to $60$ times) 
by pre-processing environmental data before optimization. 

To mitigate the inherent limitations of MPC in exploring large-scale unknown environments, 
i.e., struggling to (effectively) incorporate multi-level environmental information 
in its optimization-based decision making, we introduce 
a bi-level parent-child FLMPC architecture. The child FLMPC layer focuses on short-term, 
reactive decision making, while the parent FLMPC layer provides long-term strategic guidance for robots. 
Such a structured coordination ensures a balanced and globally (close-to) optimal exploration, 
and reduces the impact of randomness on worst-case mission time, resulting in 
enhanced reliability in search and rescue missions via robots.  
The results of the second case study show that bi-level parent-child FLMPC ensures a more balanced exploration 
and reduces randomness and delays in finding all the victims. 
While single-level FLMPC is in some setups faster, it risks inefficiencies 
and struggling to find victims in less accessible places, whereas the bi-level 
parent-child FLMPC improves consistency and reliability in mission completion.%

Future research on FLMPC should focus on optimizing computational efficiency, 
improving parameter tuning, and enhancing adaptability to dynamic environments. 
Developing specialized software for fuzzy optimization and integrating heuristic algorithms 
will reduce the computation time and will enhance the optimality. 
A structured or data-driven approach to parameter tuning, in absence of 
data scarcity and randomness in data, will eliminate reliance on trial and error. 
Additionally, making the bi-level parent-child FLMPC framework more dynamic 
by refining the procedure of fuzzy cluster formation, 
enabling real-time cell re-allocation, and adapting the scoring 
method for the fuzzy clusters will enhance efficiency and adaptability 
of bi-level parent-child FLMPC. 
In particular, the limitation discussed in Remark~\ref{rem:limitation_bilevel_FLMPC}, 
i.e., premature association of the state ``explored'' to certain fuzzy clusters 
due to low detectability of high-reward cells, should be addressed 
to further enhance the robustness and reliability of the bi-level parent-child FLMPC system. 
It is hypothesized that careful tuning of the parameters (e.g., $\sexp$ and $w^\text{score}$)  
will address the issue by compromising the mission speed. This hypothesis should be evaluated via extensive simulations.  
The impact of a more dynamic approach, considering flexibility in the shape 
of fuzzy clusters, on improving the area coverage and effectiveness of 
the bi-level parent-child FLMPC in finding all the victims should also be investigated. 
Finally, implementing the bi-level parent-child FLMPC system in realistic 
search and rescue operations, including dynamic environments, multiple tasks, 
and diverse robots should be researched.%


\begin{thebibliography}{00}


{

\bibitem[1]{disaster_website}
CRED, Em-dat the international disaster database. http://www.emdat.be/, 2025 (accessed 13 February 2025).

\bibitem[2]{LocationNavigationSurvey}
C. Fischer, H. Gellersen 
 Location and navigation support for emergency responders: A survey,
 {IEEE Pervasive Computing}. 9 (2010) 38--47.  https://doi.org/10.1109/MPRV.2009.91

 \bibitem[3]{longMissionTimes}
S.F. Ochoa, R. Santos, Human-centric wireless sensor
networks to improve information availability during urban search and
rescue activities, Information Fusion, 22 (2015) 71--84. https://doi.org/10.1016/j.inffus.2013.05.009

\bibitem[4]{humanRobotTeam}
 A. Bock, {\AA}. Svensson, A. Kleiner, J. Lundberg, T. Ropinski 
 A visualization‐based analysis system for urban search \& rescue mission planning support,
 {Computer Graphics Forum}. 36 (2017) 148--159.  https://doi.org/10.1111/cgf.12869

  \bibitem[5]{Fukushima}
CBS News,  Robots come to the rescue after Fukushima Daiichi nuclear disaster. https://www.cbsnews.com/news/robots-fukushima-daiichi-nuclear-disaster-60-minutes-2021-07-11/, 2021 (accessed 08 February 2025).

\bibitem[6]{UAV_USARsurvey}
R. Zurli, R. Bravo, A. Leiras 
 Literature review of the applications of UAVs in humanitarian relief, XXXV Encontro Nacional de Engenharia de Produção. (2015)



 \bibitem[7]{icarus}
G. De~Cubber, D. Doroftei, Y. Baudoin, D. Serrano, K. Berns, C. Armbrust, K. Chintamani, R. Sabino, S. Ourevitch, T. Flamma, 
 Search and rescue robots developed by the European Icarus project, {7th IARP workshop on ‘Robotics for Risky Environment and Extreme Robotics’}. (2013).

 \bibitem[8]{snakeRobotsMexico}
J. Whitman, N. Zevallos, M. Travers, H. Choset, 
 Snake robot urban search after the 2017 Mexico city earthquake,
 In {2018 IEEE International Symposium on Safety, Security, and Rescue Robotics (SSRR)}. (2018) 1--6. https://doi.org/10.1109/SSRR.2018.8468633

  \bibitem[9]{darpa}
M. Tranzatto, T. Miki, M. Dharmadhikari, L. Bernreiter, M. Kulkarni, F. Mascarich, O. Andersson, S.M.K Khattak, M. Hutter, R. Siegwart, K. Alexis,
 CERBERUS in the DARPA Subterranean Challenge,
{Science Robotics}. 7  (2022). https://doi.org/10.1126/scirobotics.abp9742

 \bibitem[10]{fuzzyMPCsearch}
de~Koning C, Jamshidnejad A. 
 Hierarchical integration of model predictive and fuzzy logic control for combined coverage and target-oriented search and rescue via robots with imperfect sensors,
 {Journal of Intelligent {\&} Robotic Systems}. 40 (2023). https://doi.org/10.1007/s10846-023-01833-2

\bibitem[11]{MPCbook}
J. Rawlings, D.Q. Mayne, M. Diehl,
 {Model Predictive Control: Theory, Computation, and Design}, second ed.,
Madison, USA, 2017.

\bibitem[12]{motionEncodedPS}
M.D Phung, Q.P. Ha  
 Motion-encoded particle swarm optimization for moving target search using UAVs,
 {Applied Soft Computing}. 97 (2020). https://doi.org/10.1016/j.asoc.2020.106705


\bibitem[13]{priorityEncoding}
Z. Wang, J. Guo, W. Zou, S. Li, 
 Cooperative search for moving targets with the ability to perceive and evade using multiple UAVs,
 {Intelligence \& Robotics}. 3 (2023) 538--564. https://doi.org/10.20517/ir.2023.30
 

\bibitem[14]{Fuzzy_introduction}
L.A. Zadeh,  
 Fuzzy sets,
 {Information and Control}. 8(3) (1965) 338--353. https://doi.org/10.1016/S0019-9958(65)90241-X

 \bibitem[15]{bel}
R.E. Bellman, L.A. Zadeh  
 Decision-making in a fuzzy environment,
 {Management Science}. 17 (1970) 141--164. https://doi.org/10.1287/mnsc.17.4.B141

 \bibitem[16]{fuzzy_optimization_aggregation_property}
R.R. Yager,  
 On ordered weighted averaging aggregation operators in multicriteria decisionmaking,
 {IEEE Transactions on Systems, Man, and Cybernetics}. 18(1) (1988) 183--190. https://doi.org/10.1109/21.87068

  \bibitem[17]{zimmermanFuzzyProgramming}
H.J. Zimmermann 
 Fuzzy programming and linear programming with several objective functions,
 {Fuzzy Sets and Systems}. 1 (1978) 45--55. https://doi.org/10.1016/0165-0114(78)90031-3

  \bibitem[18]{fuzzyMPC}
JM. da~Costa~Sousa, U. Kaymak  
 Model predictive control using fuzzy decision functions,
 {IEEE Transactions on Systems. Man, and Cybernetics, Part B (Cybernetics)}, 31(1) (2001) 54--65.

\bibitem[19]{fuzzyMPCstability}
K. Belarbi, F. Megri (2007)
A stable model-based fuzzy predictive control based on fuzzy dynamic programming, 
 {Fuzzy Systems, IEEE Transactions on}. 15 (2007) 746 -- 754. https://doi.org/10.1109/TFUZZ.2006.890656

 \bibitem[20]{fuzzyMPCstabilityExtended}
T. Samir, K. Belarbi, 
 A stabilizing model fuzzy predictive control scheme for nonlinear deterministic systems,
 IEEE International Conference on Fuzzy Systems. 07 (2016) 724--729. https://doi.org/10.1109/FUZZ-IEEE.2016.7737759.

\bibitem[21]{Surma2024}
F. Surma, A. Jamshidnejad, 
State-dependent dynamic tube MPC: A novel tube MPC method with a fuzzy model of disturbances, 
{International Journal of Robust and Nonlinear Control}. 36 (2024) . https://doi.org/10.1002/rnc.7558

  \bibitem[22]{fuzzyModels}
D. Shen, C.C. Lim, P. Shi, 
 Robust fuzzy model predictive control for energy management systems in fuel cell vehicles,
{Control Engineering Practice}. 98 (2020). https://doi.org/10.1016/j.conengprac.2020.104364

  \bibitem[23]{FuzzyCooperativeSecondaryControl}
Y. Shan, J. Hu, K. Chan, S. Islam,  
 A Unified Model Predictive Voltage and Current Control for Microgrids With Distributed Fuzzy Cooperative Secondary Control,
{IEEE Transactions on Industrial Informatics}. 17 (2021) 8024-8034. https://doi.org/10.1109/TII.2021.3063282

\bibitem[24]{Baglioni2024}
M. Baglioni, A. Jamshidnejad 
 A novel MPC formulation for dynamic target tracking with increased area coverage for search and rescue robots, 
{Journal of Intelligent and Robotic Systems}. 110 (2024). https://doi.org/10.1007/s10846-024-02167-3

\bibitem[25]{Surma2024_IFAC}
F. Surma, A. Jamshidnejad, 
 Approximate SDD-TMPC with spiking neural networks: An application to wheeled robots, 
{IFAC-Papers Online}. 58 (2024)  323-328. https://doi.org/10.1016/j.ifacol.2024.09.050


\bibitem[26]{motionEncodedGA}
M. Alanezi, H. Bouchekara, T. Apalara, M. Shahriar, Y. Shaaban, M. Javaid, M. Khodja 
 (2022)  Dynamic target search using multi-UAVs based on motion-encoded genetic algorithm with multiple parents,
{ IEEE Access}. 10 (2022) 77922-77939. https://doi.org/10.1109/ACCESS.2022.3190395

 \bibitem[27]{distrbuitedCooperativeSearch}
H. Zhang, H. Ma,  Mersha BW, Zhang X, and Jin Y 
 Distributed cooperative search method for multi-UAV with unstable communications,
 {Applied Soft Computing}. 148 (2023). https://doi.org/10.1016/j.asoc.2023.110592

\bibitem[28]{searchUAVlearning}
Y. Yang, M. Polycarpou, A. Minai, 
 Multi-UAV cooperative search using an opportunistic learning method,
 {Journal of Dynamic Systems Measurement and Control-transactions of The Asme - J DYN SYST MEAS CONTR}. 129 (2007).  https://doi.org/10.1115/1.2764515

\bibitem[29]{ProbabiliticRobotics}
S. Thrun, W. Burgard, D. Fox, 
 {Probabilistic Robotics (Intelligent Robotics and Autonomous Agents)}, first ed.,
 Massachusetts, USA, 2005.

\bibitem[30]{motionEncodedSCS}
Y. Niu, X. Yan, Y. Wang, Y. Niu, 
An improved sand cat swarm optimization for moving target search by UAV,
 {Expert Systems with Applications}. 238 (2024). https://doi.org/10.1016/j.eswa.2023.122189

 \bibitem[31]{DMPCsearch}
P. Trodden, A. Richards, 
 Multi-vehicle cooperative search using distributed model predictive control,
 {AIAA Guidance, Navigation and Control Conference and Exhibit}. (2008). https://doi.org/10.2514/6.2008-7138

\bibitem[32]{antColony}
S. Perez-Carabaza, E. Besada-Portas, JA. Lopez-Orozco, and de~la Cruz JM. 
 Ant colony optimization for multi-UAV minimum time search in uncertain domains,
 {Applied Soft Computing}. 62 (2017) 789-806. https://doi.org/10.1016/j.asoc.2017.09.009

\bibitem[33]{BayesianOptimizationAglortihm}
 P. Lanillos, J. Yañez-Zuluaga, J. Ruz, E. Besada, 
 A bayesian approach for constrained multi-agent minimum time search in uncertain dynamic domains,
 {Gecco'13: Proceedings of the 2013 Genetic and Evolutionary Computation Conference}. 7 (2013) 391--398. https://doi.org/10.1145/2463372.2463417

 \bibitem[34]{CEE:2024}
A. Jamshidnejad, B. De Schutter, 
A combined probabilistic-fuzzy approach for dynamic modeling of traffic in smart cities: Handling imprecise and uncertain traffic data, 
{Computers and Electrical Engineering}. 119 (2024) . https://doi.org/10.1016/j.compeleceng.2024.109552

\bibitem[35]{fuzzyControl}
K.M. Passino, S. Yurkovich,
 {Fuzzy Control}, first ed.,
 Longman, USA, 1997.



\bibitem[36]{Zadeh:M342}
L.A. Zadeh, 
 Outline of a new approach to the analysis of complex systems and decision processes,
 Transactions on Systems, Man, and Cybernetics. 1 (1972) 28-44. https://doi.org/10.1109/TSMC.1973.5408575

 \bibitem[37]{serialParallel}
RR. Negenborn, B.  {De Schutter} , J. Hellendoorn  
 Multi-agent model predictive control for transportation networks: Serial versus parallel schemes,
{Engineering Applications of Artificial Intelligence}. 21 (2008) 353-366. https://doi.org/10.1016/j.engappai.2007.08.005

\bibitem[38]{BiLevel}
A. Jamshidnejad, D. Sun, A. Ferrara, B. De Schutter 
 A novel bi-level temporally-distributed MPC approach: An application to green urban mobility,
{Transportation Research Part C: Emerging Technologies}. 156 (2023). https://doi.org/10.1016/j.trc.2023.104334 

\bibitem[39]{MatlabPS}
Particleswarm Mathworks https://www.mathworks.com/help/gads/particleswarm.html. Accessed 08 February 2025   

\bibitem[40]{ParticleSwarm}
J. Kennedy, R. Eberhart 
 Particle swarm optimization,
 In {Proceedings of ICNN'95 - International Conference on Neural Networks}. 4 (1995) 1942--1948. https://doi.org/10.1109/ICNN.1995.488968.

  \bibitem[41]{codeFLMPC}
F. Surma, Implementation of FLMPC v1.0 [software], 4TU.ResearchData, February 18, 2025 https://doi.org/10.4121/319168f0-bc62-4051-84c2-f32718c05386.

 \bibitem[42]{delftBlue}
TU Delft, {{D}elft {H}igh {P}erformance {C}omputing {C}entre ({DHPC})}
 {{D}elft{B}lue {S}upercomputer ({P}hase 2)}. https://www.tudelft.nl/dhpc/ark:/44463/DelftBluePhase2, 2024 (Accessed 08 February 2025).

\bibitem[43]{VRP}
S. Nanda~Kumar, R. Panneerselvam, 
 A survey on the vehicle routing problem and its variants,
 {Intelligent Information Management}. (2012). https://doi.org/10.4236/iim.2012.43010

\bibitem[44]{mTSP}
O. Cheikhrouhou, I. Khoufi 
 A comprehensive survey on the multiple traveling salesman problem: Applications, approaches and taxonomy,
 {Computer Science Review}. 40 (2021). https://doi.org/10.1016/j.cosrev.2021.100369



\bibitem[45]{mTSPga}
A. Kir{\'a}ly, J. Abonyi 
 {A Novel Approach to Solve Multiple Traveling Salesmen Problem by Genetic Algorithm},  
 Springer Berlin Heidelberg, Berlin, Heidelberg. 313 (2010) 141--151. https://doi.org/10.1007/978-3-642-15220-7\_12



\bibitem[46]{mTSPcode}
A. Király , Multiple Traveling Salesmen Problem - Genetic Algorithm, using multi-chromosome representation v1.11 [software], {mTSPcode}, (2024) https://www.mathworks.com/matlabcentral/fileexchange/48133-multiple-traveling-salesmen-problem-genetic-algorithm-using-multi-chromosome-representation.

}
\end{thebibliography}
\end{document}